\documentclass[a4paper,fleqn]{cas-sc}

\usepackage[authoryear]{natbib}

\usepackage{multicol} 

\usepackage{times}
\usepackage{algorithmic}
\usepackage{algorithm}
\usepackage{latexsym}
\usepackage{amssymb}
\usepackage{amsmath}
\usepackage{amsthm}
\usepackage{booktabs}
\usepackage{enumitem}
\usepackage{graphics} 
\usepackage{epsfig} 
\usepackage{kotex}
\usepackage{xcolor}

\usepackage{colortbl} 
\usepackage{graphicx} 
\usepackage{multirow} 
\usepackage{xcolor} 
\usepackage{subcaption}
\usepackage{caption}

\usepackage{hyperref}

\def\tsc#1{\csdef{#1}{\textsc{\lowercase{#1}}\xspace}}
\tsc{WGM}
\tsc{QE}

\definecolor{lightgreen}{RGB}{157,222,124}
\definecolor{yellow}{RGB}{255,213,93}
\definecolor{blue_arrow}{RGB}{68,114,193}
\definecolor{purple_arrow}{RGB}{112,48,160}

\begin{document}
\let\WriteBookmarks\relax
\def\floatpagepagefraction{1}
\def\textpagefraction{.001}

\shorttitle{}    

\shortauthors{}  

\title [mode = title]{Foundation Model-Driven Framework for Human-Object Interaction Prediction with Segmentation Mask Integration}  

\tnotemark[1] 


%
\author[1]{Juhan Park}[orcid=0009-0004-1105-9048]
\ead{p3549823@cau.ac.kr}
\credit{Conceptualization, Data curation, Formal analysis, Investigation, Methodology, Validation, Writing - original\&draft}
\affiliation[1]{organization={Department of Artificial Intelligence, Chung-Ang University},
            addressline={84 Heukseok-ro, Dongjak-gu}, 
            city={Seoul},
            postcode={06974}, 
            country={Republic of Korea}}

\author[2]{Kyungjae Lee}[orcid=0000-0003-0147-2715]
\cormark[1]
\ead{kyungjae_lee@korea.ac.kr}
\credit{Conceptualization, Supervision, Writing - review \& editing}
\affiliation[2]{organization={Department of Statistics, Korea University},
            addressline={145 Anam-ro, Seongbuk-gu}, 
            city={Seoul},
            postcode={02841}, 
            country={Republic of Korea}}
            
\author[3]{Hyung Jin Chang}[orcid=0000-0001-7495-9677]
\ead{h.j.chang@bham.ac.uk}
\credit{Supervision, Writing - review \& editing}
\affiliation[3]{organization={School of Computer Science, University of Birmingham},
            addressline={Edgbaston}, 
            city={Birmingham},
            postcode={B15 2TT}, 
            country={United Kingdom}}
            
\author[4]{Jungchan Cho}[orcid=0000-0002-3859-1702]
\cormark[2]
\ead{thinkai@gachon.ac.kr}
\credit{Conceptualization, Funding acquisition, Project administration, Supervision, Writing - review \& editing}
\affiliation[4]{organization={School of Computing, Gachon University},
            addressline={1342 Seongnamdaero}, 
            city={Sujeong-gu},
            state={Gyeonggi-do},
            postcode={13120}, 
            country={Republic of Korea}}

\cortext[1]{Corresponding author. Department of Statistics, Korea University, 145 Anam-ro, Seongbuk-gu, Seoul, 02841, Republic of Korea.}
\cortext[2]{Corresponding author. School of Computing, Gachon University, 1342 Seongnamdaero, Sujeong-gu, Gyeonggi-do, 13120, Republic of Korea}


\begin{abstract}
In this work, we introduce Segmentation to Human-Object Interaction (\textit{\textbf{Seg2HOI}}) approach, a novel framework that integrates segmentation-based vision foundation models with the human-object interaction task, distinguished from traditional detection-based Human-Object Interaction (HOI) methods.
Our approach enhances HOI detection by not only predicting the standard triplets but also introducing quadruplets, which extend HOI triplets by including segmentation masks for human-object pairs. 
More specifically, Seg2HOI inherits the properties of the vision foundation model (e.g., promptable and interactive mechanisms) and incorporates a decoder that applies these attributes to HOI task. 
Despite training only for HOI, without additional training mechanisms for these properties, the framework demonstrates that such features still operate efficiently. 
Extensive experiments on two public benchmark datasets demonstrate that Seg2HOI achieves performance comparable to state-of-the-art methods, even in zero-shot scenarios. 
Lastly, we propose that Seg2HOI can generate HOI quadruplets and interactive HOI segmentation from novel text and visual prompts that were not used during training, making it versatile for a wide range of applications by leveraging this flexibility.
\end{abstract}




\maketitle

\section{Introduction}
Understanding human-environment interaction is crucial for building effective interactive systems, as it allows machines to interpret and respond to human behaviors in a natural and context-aware manner.
A prime example of such a system is generative pre-trained transformers 4 (GPT-4) \citep{achiam2023gpt} which excels at linguistic understanding.
However, for many real-world applications, such as collaborative systems or robotics, linguistic understanding alone is not sufficient. 
Accurate decision-making requires integrating both linguistic and visual comprehension to fully capture the context in which interactions occur.
Visual understanding is essential for tasks such as recognizing objects, interpreting gestures, and understanding spatial relationships.
Without this capability, interactive systems lack the necessary depth to function effectively in dynamic environments.
\begin{figure}
	\centering
    \begin{subfigure}[b]{0.225\textwidth}
        \centering
        \includegraphics[width=\textwidth]{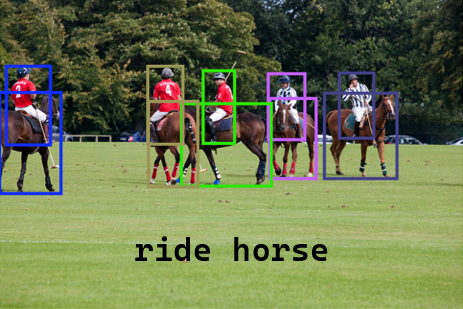}
        \caption{HOI triplet}\label{fig:hoi_triplet}
    \end{subfigure} \hspace{0.005\linewidth}
    \begin{subfigure}[b]{0.225\textwidth}
        \centering
        \includegraphics[width=\textwidth]{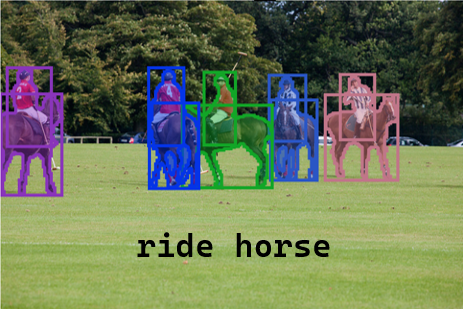}
        \caption{HOI quadruplet}\label{fig:hoi_quadruplet}
    \end{subfigure}  \hspace{0.02\linewidth}
    \begin{subfigure}[b]{0.47\textwidth}
        \centering
        \includegraphics[width=\textwidth]{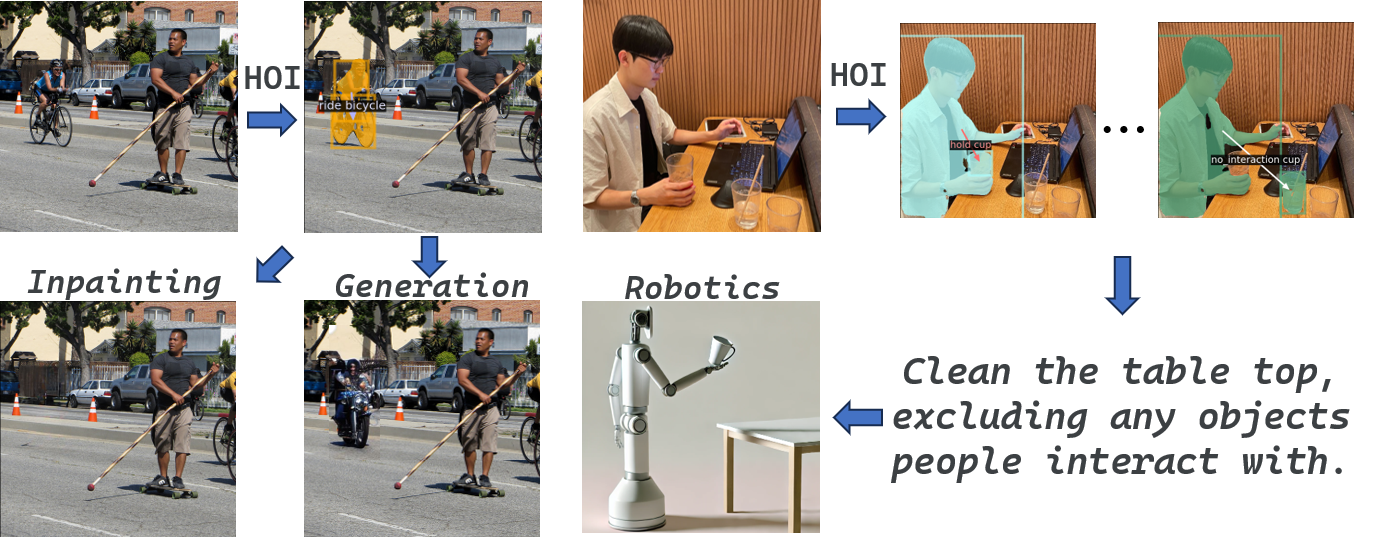}
        \caption{Interactive System Application}\label{fig:hoi_application}
    \end{subfigure}
	\caption{{In (a), the prediction set from traditional HOI detection methods, while (b) illustrates the proposed method from this paper, which includes HOI segmentation in addition to standard prediction. In (c), examples of potential vision and robotic applications are presented.}}
	\label{FIG:0}
    \vspace{-5pt}
\end{figure}

In this regard, interactive foundation models are increasingly being applied to intelligent interactive systems where accurate decision-making depends not only on linguistic judgment but also on visual understanding. To support this, many researchers in the field of computer vision have focused on human-object interactions (HOIs) \citep{chao2018learning}.
However, traditional HOI approaches, as shown in Fig. \ref{fig:hoi_triplet}, are limited to identifying triplets of <human, object, action>, which restricts their applicability to broader fields. We propose a new concept of quadruplet <human, object, action, union-mask>, as illustrated in Fig. \ref{fig:hoi_quadruplet}, which includes a union segmentation to capture the fined-interaction region between human and object.
This new quadruplet significantly enhances its potential for diverse applications. For instance, in collaborative robotics with humans, as depicted in Fig.~\ref{fig:hoi_application}, the robot not only needs to perform accurate instance-level depth segmentation but also must understand the interactions between humans and objects.

The main challenge in achieving this lies in generating accurate ground truth data for union masks. 
Current research trends heavily rely on traditional supervised learning methods, where HOI data is collected separately, and models are retrained for each specific task.
This approach is not only resource-intensive but also impractical when dealing with complex interactions like those requiring union masks. 
Moreover, the lack of accurate mask annotations for human-object pairs further complicates the development of effective models.
As artificial intelligence (AI) models continue to scale, HOI research should shift towards more efficient approaches by leveraging vision foundation models. 
To overcome these limitations and achieve the proposed quadruplet, this paper utilizes segmentation-based vision foundation models, which are pretrained on large datasets, and incorporates it with HOI downstream tasks. 
This approach facilitates more efficient resource utilization and advances the understanding of human-object interactions.

The Segment anything model (SAM) is a notable foundation model in computer vision, which has shown that segmentation tasks can serve as a foundation for various downstream tasks~\citep{kirillov2023segment, ravi2024sam}.
However, the segmentation task mainly focuses on pixel-level information, emphasizing low-level features, while HOI tasks require high-level features. Accordingly, combining segmentation and HOI tasks to create a unified model is very challenging. 
There has been an attempt to solve segmentation and HOI simultaneously \citep{zhou2020cascaded}. However, this approach has limitations as it relies on joint training of a multi-task model using both ground truth human-object pairs and segmentation masks. It fundamentally differs from what we are pursuing here, which is to preserve the original tasks of the foundation model while adding new tasks without the need for retraining the entire model. \
This allows for more efficient utilization and seamless integration of new tasks without compromising the performance of the existing model.
To the best of our knowledge, existing research does not handle the HOI task based on a segmentation foundation model. \emph{How to leverage the foundation model to aid in the learning of high-level features for other tasks remains an open question.}

In this paper, we focus on freezing a pretrained segmentation foundation model and attaching a learnable HOI-specific task decoder in a cascade manner. 
This approach aims to preserve the strengths of the foundation model in handling low-level features while effectively adapting it to perform high-level tasks such as HOI detection. 
Fig.~\ref{fig:overview} depicts the proposed novel unified segmentation-based HOI detection framework, named \textbf{Seg2HOI}, which consists of a foundation model for the segmentation task and an HOI decoder.
The HOI decoder learns relation features from instance features produced by the vision decoder.
Unlike the common two-stage HOI approach, which requires explicit human-object pair matching to simplify implementation, we utilize implicit cross-attention blocks between human and object features. This design maintains a consistent number of queries for the model to handle, even as the number of detections in an image increases. 
These advantages are expected to enable the expansion not only to HOI tasks but also to Object-Object Interaction (OOI) \citep{unmesh2023interacting} with minimal additional cost.

Also, as a novel approach in quadruplet training, our method directly incorporates the learning of union mask and intersection mask information for human-object pairs within the HOI decoder. This contributes to improving HOI performance without the need for additional foundation model training. One of the challenges in training HOI segmentation is the lack of ground truth mask data in existing HOI datasets, as they are detection-based. To address this issue, we propose generating pseudo-labels for HOI masks by utilizing instance masks obtained by a segmentation-based foundation model, facilitating the learning of HOI masks in a self-supervised manner. 
Furthermore, we demonstrated the extensibility of the proposed Seg2HOI to various backbones \citep{zou2024segment} by expanding our method to interactive segmentation using visual and text prompt inputs. We also proposed interactive human-object image segmentation using the quadruplet approach. As shown in Fig. \ref{fig:hoi_application}, this interactive HOI segmentation, combined with inpainting, can give rise to new and useful applications.

In our experiments, conducted on the Humans Interacting with Common Objects Detection (HICO-DET) \citep{chao2018learning} and Verbs in Common Objects in Context (V-COCO) \citep{gupta2015visual} datasets, we demonstrate that Seg2HOI achieves comparable performance to existing state-of-the-art methods based on triplets, even when solving the quadruplet problem with the addition of HOI segmentation. This holds true not only for the closed-set vocabulary HOI problem but also for the open-set vocabulary HOI problem. Additionally, we showcase the potential of the newly proposed interactive HOI image segmentation, and finally, we present an ablation study analyzing the impact of mask addition on HOI learning.

Our contributions can be summarized as follows: 
\begin{itemize} 
\item We integrate vision foundation models with human-object interaction (HOI), achieving performance comparable to state-of-the-art methods without compromising the foundation models' performance. To accomplish this, we propose implicit human-object relation learning and multi-head HOI prediction.
\item Next, we define and solve the quadruplet problem in human-object interaction task, including HOI segmentation. 
To address the lack of mask data in existing HOI datasets, we propose HOI mask pseudo labeling, enabling training for HOI segmentation and leading to significant performance improvements.
\item Additionally, our approach, Seg2HOI, demonstrates performance comparable to state-of-the-art methods in zero-shot HOI tasks without requiring additional loss mechanisms for open-vocabulary handling.
\item Finally, we extend our work to include interactive human-object interaction image segmentation. 
This extension enhances the practical applications of our approach by effectively supporting interactive downstream HOI task.
\end{itemize}

\section{Related Works}
\subsection{Object detection and Segmentation}
Existing HOI models approach the problem as an extension of object detection (OD). Thus, it is essential to review the evolution of object detection and segmentation (Seg.) fields. Initially, classical convolutional neural network (CNN)-based models were developed to train object detection and segmentation separately \citep{ren2015faster, he2017mask, long2015fully}. Subsequently, unified models that combined different tasks emerged, demonstrating improved performance \citep{chen2019hybrid, zhang2021k}.  Recently, the advent of Transformer-based models \citep{carion2020end, zhang2022dino, liu2022dab, li2022dn, cheng2021per, cheng2022masked} has brought significant advancements in the fields of detection and segmentation. 
Additionally, Transformer-based unified models, such as Mask-DINO \citep{li2023mask}, K-Net \citep{zhang2021k}, OpenSEED \citep{zhang2023simple}, FreeSeg \citep{qin2023freeseg}, BURST \citep{athar2023burst}, and UniVS \citep{li2024univs}, have been introduced, allowing for a broader range of tasks—including object detection, instance segmentation (Ins. Seg.), and panoptic segmentation—to be handled by a single model, thereby contributing to performance improvements. Similarly, there is a growing body of research focused on enhancing performance and efficiency by consolidating multiple tasks within a single framework. 

Given this trend, while C-HOI \citep{zhou2020cascaded} can handle OD-HOI and Seg-HOI configurations, it is not well-suited for integrating OD, Seg., and HOI into a unified framework. Additionally, as a CNN-based approach, it lags behind transformer-based architectures in terms of performance and scalability. In this study, we aim to extend the conventional HOI task. Following the discussion on Mask-DINO, by providing pixel-level detailed guidance to the HOI decoder, it is possible to extend HOI models in line with the performance improvements seen in unified models. This is achieved in our approach by directly guiding the HOI decoder with HOI segmentation, such as the union and intersection masks of the human-object pairs.

\begin{table}[t]
    \centering
    \caption{Comparison of different methods. Proposed V operates solely based on visual features, while Proposed I functions interactively, accepting various prompts and supporting open vocabulary. OD refers to object detection. HOI stands for human-object interaction. Interactive indicates whether the model supports interaction with the user, allowing it to receive various prompts or adapt to user input dynamically.}
    \begin{tabular}{lcccccc}
        \toprule
        Method & OD & HOI & Interactive & Ins. Seg. & HOI Seg. \\
        \midrule
        Mask-DINO \citep{li2023mask} & \checkmark &   & & \checkmark & \\
        SEEM \citep{zou2024segment} & \checkmark & & \checkmark & \checkmark & \\
        \midrule
        Proposed V (Sec. \ref{sec:seg2hoi}) & \checkmark & \checkmark & & \checkmark & \checkmark \\
        Proposed I (Sec. \ref{sec:seem2hoi}) & \checkmark & \checkmark & \checkmark & \checkmark & \checkmark \\
        \bottomrule
    \end{tabular} 
    \label{table:1}
\end{table}

\subsection{Human-object interaction recognition}
Formally, HOI involves detecting human-object pairs and identifying the corresponding interaction verbs simultaneously, such as \(\langle \text{human}, \text{object}, \text{action} \rangle\) triplets.
Existing HOI detectors can be categorized into one-stage methods and two-stage methods. Both have advanced significantly with deep learning, but each has its drawbacks.
The one-stage method \citep{chen2021reformulating, fang2021dirv,  zou2021end, zhong2021glance, Liao_2022_CVPR, NEURIPS2021_8f1d4362, Kim_2023_CVPR} addresses HOI detection by using multiple heads to solve human-object pairs and interacting verbs with a single query in an end-to-end manner. Despite its streamlined approach, this method has the disadvantage of being specialized solely for HOI tasks. 
On the other hand, the two-stage method \citep{gkioxari2018detecting, gao2018ican, hou2020visual, ulutan2020vsgnet, li2020pastanet, kim2021hotr, zhang2022exploring, zhang2023exploring} first detects objects and humans, and then arbitrarily pairs them before classifying the interaction. However, this approach struggles with the complexity of post-processing and the need for additional matching modules, highlighting the inherent complications and limitations of this method.
From an application perspective, adding new tasks and retraining the entire foundation model are inefficient. Furthermore, operating two separate models to perform both segmentation and HOI tasks is highly impractical and resource-intensive.
Our work adopts a two-stage approach but simplifies the process by learning HOI directly without the need for an additional pairing process following the pretrained detector. 

\subsection{Various application in computer vision tasks}
Foundation models in various fields output general features, making them widely used in a variety of downstream tasks \citep{carion2020end, dosovitskiy2020image, radford2021learning, li2023blip, rombach2022high}. Recently, various general research studies have emerged that accept text and visual prompts in a free-form manner. By demonstrating few-shot or zero-shot performance, these models offer the advantage of being applicable in any environment. In the realm of Large Language Models, GPT exemplifies this approach, while SAM serves as a representative foundation model for vision tasks. Applications are also diverse, including VisionLLM \citep{wang2024visionllm}, InternGPT\citep{liu2023interngpt}, SEEM\citep{zou2024segment}, and LLaVA interactive\citep{chen2023llava}, all of which enable interactive communication with human users. Additionally, HOI4ABOT \citep{mascaro2023hoi4abot} demos are emerging in human-robot interaction, showcasing the potential for interactive applications beyond just human users. These demos operate not only based on text but also from various human preferences such as points, boxes, and referred regions. In this paper, we extend these capabilities to the HOI task, demonstrating the versatility of HOI segmentation in various applications when receiving text or visual prompts.

\section{Backgrounds: Vision Foundation Model}\label{sec:background}
The framework proposed in this paper aims to maintain the original task of the foundation model while adding HOI capabilities. Accordingly, we adopt a two-stage pipeline to integrate HOI functionality. In Sec.~\ref{sec:trans} and \ref{sec:hung}, we describe the foundation model for fundamental tasks (e.g. object detection, segmentation) and training processes, respectively. In Sec.~\ref{sec:prob_def}, we explain the problem definition and the direction in which we aim to solve it. In Sec.~\ref{sec:seg2hoi}, we introduce our segmentation-based HOI approach.

\subsection{Transformer-Based Framework}\label{sec:trans}
\begin{figure*}
    \centering
    \includegraphics[width=1\textwidth]{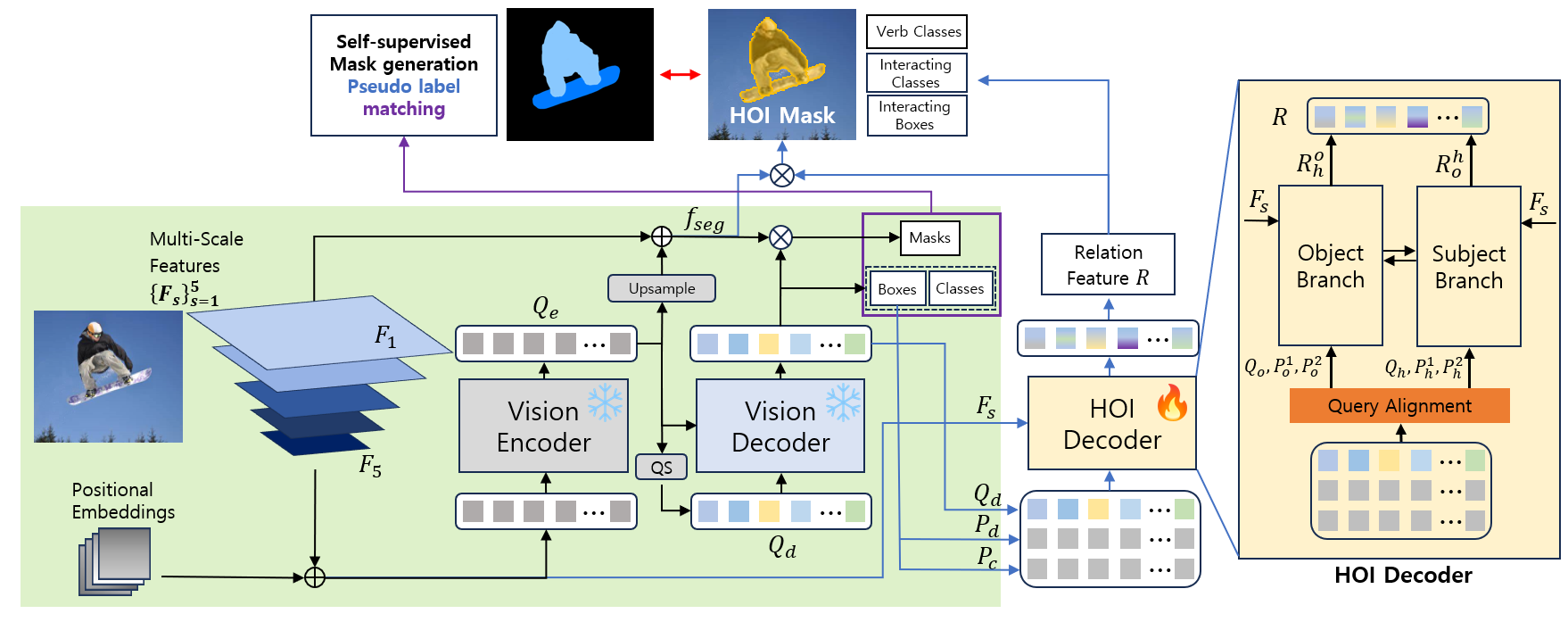}
    \vspace{-12pt}
    \caption{Overview of the proposed Seg2HOI framework which consists of two steps: \textcolor{lightgreen}{pretrained vision foundation model} and \textcolor{yellow}{HOI model}. The black arrow represents the process of the pretrained segmentation model, the \textcolor{blue_arrow}{blue arrow} represents the process added in this paper, and the \textcolor{purple_arrow}{purple arrow} represents the process for pseudo-labeling.}
    \label{fig:overview}
    \vspace{-10pt}
\end{figure*}

The proposed method can leverage various segmentation-based foundation models. In this paper, we adopt the foundation model \citep{li2023mask} highlighted in the green background of Fig.~\ref{fig:overview}, which offers both segmentation and object detection capabilities for the HOI downstream task.

Given an image $X \in \mathbb{R}^{H \times W \times 3}$, the backbone network maps the RGB input into multi-scale features $\{F_s\}_{s=1}^{S} = \text{Backbone}(X)$, where $F_{s} \in \mathbb{R}^{\frac{H}{4s} \times \frac{W}{4s} \times C_s}$ represents the feature at the $s$-th level of the pyramid and $S$ is the total number of scales. Additionally, a $1\times 1$ convolution is applied to project each $F_s$ into $F_d \in \mathbb{R}^{\frac{H}{4s} \times \frac{W}{4s} \times C_d}$, where the scale $C_s$ at each level is projected to $C_d$. Then, the multi-scale feature maps are flattened and concatenated to form \( Q_s \in \mathbb{R}^{N_s \times C_d} \), where \( N_s = \sum_{s=s_i}^{s_f} {\frac{H}{4s}} \times \frac{W}{4s} \). Here, \( s_i \) and \( s_f \) represent the starting and ending scale levels considered during the flattening process. The transformer-based vision encoder generates a richer feature map for detection by utilizing the feature map with the flattened and reduced dimension \( Q_s \) along with the positional information \( Q_{pos} \) that aids in feature learning. The vision encoder is defined as \( Q_e = f_{\text{enc}}(Q_s, Q_{pos}) \), where \( f_{\text{enc}} \) consists of multiple stacked transformer encoder layers, \( Q_e \in \mathbb{R}^{N_s \times C_d} \). 
The vision decoder employs query selection (QS) used to enhance detection performance. \citep{zhang2022dino, li2023mask, zhu2020deformable} Therefore, it leverages the top-k samples of the feature map \( Q_e \) generated by the vision encoder as input. It is defined as \( Q_d = f_{\text{dec}}(Q_e, Q_{pos}) \), where \( Q_d \in \mathbb{R}^{N_k \times C_d} \) is the content query, and \( N_k \) is the number of selected top-k samples. Here, \( f_{\text{dec}} \) consists of multiple stacked transformer decoder layers. The decoder leverages both the encoded features and positional information to refine and enhance the feature representation, ultimately facilitating more accurate detection and classification tasks. The detection heads ultimately compute the instance bounding box \( b_{inst} \in [0,1]^4 \) and the one-hot encoded instance class \( c_{inst} \in \{0,1\}^{N_{\text{inst}}} \).\footnote{To avoid confusion with the notations introduced in Sec.~\ref{sec:seg2hoi}, object-related notations $b_o$ and $c_o$ were not used.}

When obtaining segmentation masks in addition to bounding box detection, an additional process is required, thus a separate segmentation branch exists \citep{zou2024segment, li2023mask, zhang2023simple}. The pixel embedding map is used to output specific instances. This map is created by merging the $1/4$ resolution feature map \(F_1\) from the backbone with an upsampled $1/8$ resolution feature map \(Q_e\) from the vision encoder. Each query embedings $Q_d$ from the decoder is then multiplied with the pixel embedding map $f_{seg}$ to produce an output mask \(m_{inst} \in \mathbb{R}^{N_k \times \frac{H}{4} \times \frac{W}{4}}\). 
\begin{align}
f_{seg} &= \mathcal{M}(\mathcal{T}(F_1) + \textsc{upsample}(Q_e)), \label{eq/seg_head}
 \\
m_{inst} &= MLP(Q_d) \otimes f_{seg}, \label{eq/instance_mask}
\end{align} 
where $\mathcal{M}$ is the segmentation head, and $\mathcal{T}$ represents a convolutional layer that maps the channel dimension to the hidden dimension. MLP stands for Multi-Layer Perceptron :$\mathbb{R}^{C_d} \rightarrow \mathbb{R}^{C_d} $.

\subsection{Hungarian matching and loss for fundamental tasks}\label{sec:hung}
Let \( y \) be the ground truth set of fundamental tasks, and let \( \hat{y} \) denote the set of \( N_d \) outputs obtained from the pretrained detector. The actual number of object instances in an image is typically much smaller than the predicted number. Therefore, to facilitate efficient training, bipartite matching between \( y \) and \( \hat{y} \) is performed before calculating the loss. During bipartite matching, the transformer-based approaches follow the procedure outlined in DETR \citep{carion2020end} and employs the Hungarian algorithm \citep{kuhn1955hungarian}. Through this process, the optimal assignment \( \hat{\sigma} \) for all possible permutations of the \( N_d \) elements \(  \sigma \in \mathfrak{S}_{N_d} \) is obtained as 
\vspace{-5pt}
\begin{equation}
\hat{\sigma} = \underset{\sigma \in \mathfrak{S}_{N_d}}{\arg\min} \sum_{i=1}^{N} \mathcal{H}_{\text{match}}(y_i, \hat{y}_{\sigma(i)})
\label{eqs:h_match}
\end{equation}
\vspace{-15pt}
\begin{equation}
\mathcal{H}_{\text{match}}= \alpha_c\mathcal{H}^{(c)} +  \alpha_b\mathcal{H}^{(b)} + \alpha_u\mathcal{H}^{(u)} + \alpha_{e}\mathcal{H}^{(e)} + \alpha_d\mathcal{H}^{(d)}   \label{eq/fundamental_task_match_cost_function}
\end{equation}
 where the matching cost \( \mathcal{H}_{\text{match}} \) is composed of three types: the classification cost \(\mathcal{H}^{(c)}\), the box L1 cost \(\mathcal{H}^{(b)}\) \citep{ren2015faster}, the box generalized Intersection of Union (IoU) cost \(\mathcal{H}^{(u)}\) \citep{rezatofighi2019generalized}, and the mask cross-entropy cost \(\mathcal{H}^{(e)}\) and mask dice costs \(\mathcal{H}^{(d)}\) \citep{milletari2016v}. The hyper-parameters \(\alpha_c\), \(\alpha_b\), \(\alpha_u\), \(\alpha_{e}\) and \(\alpha_d\), are detailed in the equations provided in \citep{li2023mask}.

During the training phase, the total loss to be minimized is calculated for the matched pairs $(y_i, \hat{y}_{\hat{\sigma}(i)})$ as follows.
\begin{equation}\label{eq/existing_loss}
\mathcal{L}_{loss} =\lambda_c \mathcal{L}_c + \lambda_b \mathcal{L}_b + \lambda_u \mathcal{L}_u + \lambda_e \mathcal{L}_e + \lambda_d \mathcal{L}_d,
\end{equation}
where  \( \mathcal{L}_c \) is the classification loss, \( \mathcal{L}_b, \mathcal{L}_u  \) is the loss related to the bounding box, and \( \mathcal{L}_e, \mathcal{L}_d \) is the loss related to the mask. These losses include the same types of losses used in the Hungarian matching cost. The hyper-parameters \( \lambda_c \), \( \lambda_b \), \( \lambda_u \), \( \lambda_e \) and \( \lambda_d \) are used to balance each loss. 

\subsection{Problem Definition}\label{sec:prob_def}
The goal of our segmentation-based HOI detection is to predict a set of HOI instances and HOI segmentation masks. Specifically, given an image $X$, the task is to predict a set of sementation-HOI quadruplets $Z = \{m^{s}, b_{h}, (b_{o}, c_{o}), v\}$. Here, $m^{s}$ and $b_h \in \mathbb{R}^4$ represent the segmentation mask of the HOI and the bounding box of the human, $b_o \in \mathbb{R}^4$ and $c_o \in \mathbb{R}^{N_{\text{obj}}}$ represent the bounding box of the object and the object label, respectively, and $v \in \mathbb{R}^{N_{\text{verb}}}$ represents the action label, where $N_{\text{obj}}$ and $N_{\text{verb}}$ are the number of objects and interacting verbs, respectively.
Recent two-stage HOI methods attach the HOI module to a detection-based DETR-family detector without the segmentation head proposed in this paper.
As mentioned earlier, segmentation is the most fundamental task in visual information analysis, making it crucial to efficiently add additional tasks without undermining this foundational knowledge.
We have clearly highlighted the differences between existing methods and our proposed method in Table~\ref{table:1}.
Note that our proposed method has the advantage of freezing the foundation model and only adding the new HOI task, ensuring that the performance of existing tasks remains completely unaffected. 

For clarity, in the following section, we will describe the extension of HOI using the vision-only foundation model discussed in this section. We will then demonstrate the generality of our framework by extending it to incorporate interactive segmentation for HOI task in Sec.~\ref{sec:seem2hoi}.

\section{Proposed Seg2HOI Framework}\label{sec:seg2hoi}

The Seg2HOI framework, depicted in Fig.~\ref{fig:overview}, is composed of two main components: a vision foundation model and a downstream HOI decoder. 
For the vision foundation model, any pretrained segmentation model from the DETR family can be integrated.
The proposed downstream HOI decoder implicitly learns the relational information between a human and objects without the need for explicit pair generation. (Sec. \ref{subs/implicit_relation_learning})
However, despite the efficiency gained, the pretrained foundation model primarily targets low-level, pixel-specific features suited for segmentation tasks, which do not
learn the fine-grained details necessary for effective HOI understanding. To bridge this gap, we also propose multi-head prediction (Sec. \ref{subs/decoder_outputs}) and HOI mask pseudo-labeling. (Sec. \ref{subs/pseudo_label}) Finally, we describe the training and inference strategy. (Sec. \ref{subs/training})

\subsection{Implicit Human-Object Relation Learning}\label{subs/implicit_relation_learning}
Current two-stage HOI methods focus on explicitly constructing human-object pairs based on possible combinations and learning the relation information or using graphs derived from detection information to capture relations \citep{zhang2022exploring, zhang2023exploring, zhang2022efficient,  park2023viplo}. In contrast, our approach deviates from these traditional methods by not pre-constructing human-object pairs. Our method implicitly learns the relationship between queries and corresponding humans or objects within the HOI decoder, utilizing the output features $Q_d$ obtained from the vision decoder. 
As shown in Fig.~\ref{fig:overview}, the proposed HOI decoder consists of two branches: the object-based branch and the subject-based branch. Each branch constructs queries through a query alignment process using the output features from the vision decoder. Each branch implicitly learns with which counterpart the query is interacting, allowing the approach to be extended not only to HOI task but also to OOI task. Compared to methods that create pairs for all possible combinations, this approach can be more efficient as the number of queries remains fixed.

To learn the above concepts, the vision decoder's the feature $Q_d$ already contains spatial priors, allowing it to operate without additional positional embedding injection. However, due to the functional characteristics of the HOI decoder, it learns the features of the counterpart involved in the relation. During the conversion of a specific instance's query feature into a relation feature, the original feature's positional prior may be lost. Therefore, by assigning positional embedding $P_d$, the relation feature retains information about the original position of the query feature.
Typically, positional embedding is achieved using a sinusoidal function of the position \citep{vaswani2017attention} or learnable embeddings \citep{carion2020end}. Following these approaches, we also construct $P_{d} \in \mathbb{R}^{N_k \times C_d}$ from the bounding box \( b_{inst} = [c_x, c_y, w, h] \) obtained from the vision decoder's head, using sinusoidal embeddings for center positions. 
\begin{figure}
	\centering
	\includegraphics[width=.40\textwidth]{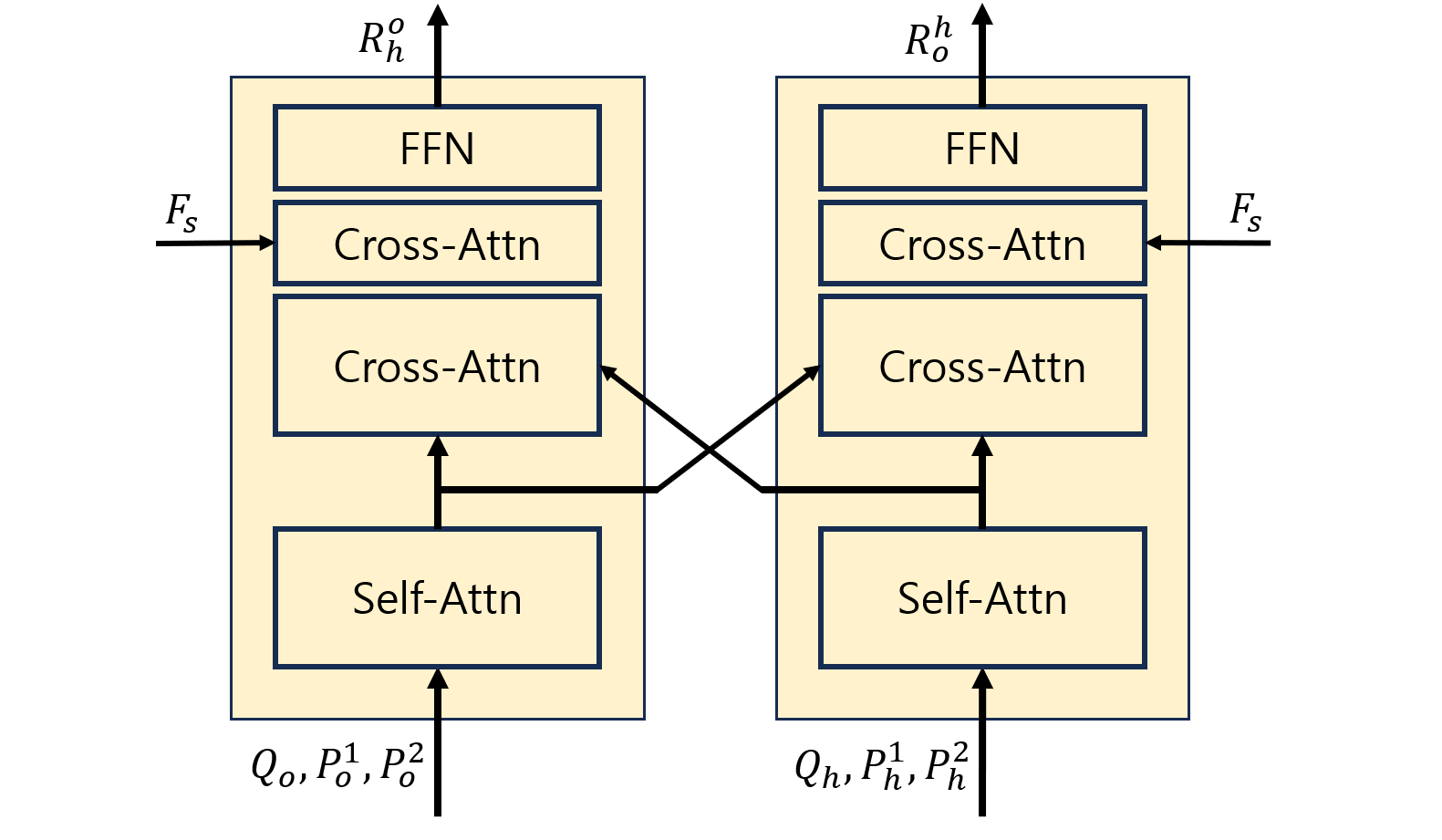}
	\caption{{Implicit pair learning in HOI deocoder.}}
	\label{figs/implicit_relation}
\end{figure}
These embeddings are concatenated with the width and height, aligned with the hidden dimensions, and used as positional embeddings for the self-attention module.
For cross-attn module within the HOI decoder, Zhang et al. \citep{zhang2023exploring} demonstrate the effectiveness of the positional embedding used by Liu et al. \citep{liu2022dab}, and our study follows this practice. The positional embedding $P_c$ can be obtained as follows: 
\begin{align}
&P_c = \text{Cat}(\phi(c_x){w_{ref}\over{w}}, \phi(c_y){h_{ref}\over{h}}) \label{eq:positional_emb_2} 
, \\ 
&w_{ref}, h_{ref} = \sigma(\text{MLP}^{\text{(ref)}}(Q_d)), 
\label{eq:ref_feature}  
\end{align} where \( \phi : \mathbb{R} \rightarrow \mathbb{R}^{C_d/2} \) is the sinusoidal function, $w_{ref}$ and $h_{ref}$ are reference value from content query using $\text{MLP}^{\text{(ref)}} : \mathbb{R}^{C_d} \rightarrow \mathbb{R}^{2}$ and $\sigma$ is the sigmoid function.

Thus, we define the queries for each branch, a process referred to as query alignment, as follows: 

\vspace{-10pt}
\begin{equation}
Q_o, P_o^1, P_o^2 = \{Q_d, P_d, P_c \mid \text{top-k of } c_{inst}\} \in \mathbb{R}^{N_o\times C_d} 
\label{eq:obj_query}
\end{equation}
\vspace{-27.5pt}
\begin{equation}
Q_h, P_h^1, P_h^2  = \{Q_d, P_d, P_c \mid c_{inst} = \text{"human"}\} \in \mathbb{R}^{N_h\times C_d}, 
\label{eq:human_query}
\end{equation}
where $Q_o, Q_h$ are content queries, $P_o$ and $P_h$ are positional embeddings corresponding for object and human, respectively, and $N_o$ and $N_h$ denote the number of object and human queries. Here, \( c_{inst} \) is the predicted class of the vision foundation model. 
The detailed process of how each branch's query learns relational information is depicted in Fig.~\ref{figs/implicit_relation}. Each branch consists of one self-attention module, two cross-attention modules and feedforward network. We first pass the query for each branch through the multi-head attention mechanism, a type of \textbf{self-attention}, to create richer features.
\vspace{-5pt}
\begin{equation}
Q_o = Q_o + F_{MHA}((Q_o + P_o^1)W_o^q, (Q_o + P_o^1)W_o^k, Q_o W_o^v), \label{eq:obj_self_attn}
\end{equation}
\vspace{-20pt}
\begin{equation}
Q_h = Q_h + F_{MHA}((Q_h + P_h^1)W_h^q, (Q_h + P_h^1)W_h^k, Q_h W_h^v), \label{eq:human_self_attn}  
\end{equation}
where \(F_{MHA}\) means the multi-head attention mechanism, \(W^q\), \(W^k\), and \(W^v\) represent the parameters for the query, key, and value, respectively. The subscripts \(o\) and \(h\) on \(W\) denote the object branch and human branch, respectively.

Second, the unary query is configured to learn information from its counterpart through the \textbf{cross-attention} module, as follows.
\vspace{-5pt}
\begin{equation}
Q_o = Q_o + F_{MHA}((Q_o + P_o^2)W_o^q, (Q_h + P_h^2)W_o^k, Q_h W_o^v),\label{eq:obj_cross_attn}
\end{equation}
\vspace{-20pt}
\begin{equation}
Q_h = Q_h + F_{MHA}((Q_h + P_h^2)W_h^q, (Q_o + P_o^2)W_h^k, Q_o W_h^v), \label{eq:human_cross_attn} 
\end{equation}

Next, to supplement the feature with global information, it undergoes refinement through the backbone feature, as empirically determined by \citep{zhang2023exploring}. Finally, the relation feature \(R\) is obtained through a feed-forward network, as shown in the following equations:
\begin{equation}
Q_o = Q_o + F_{MHA}((Q_o + P_o^2)W_o^q, F_s W_o^k, F_s W_o^v),\label{eq:obj_relation}
\end{equation}
\vspace{-25pt}
\begin{equation}
Q_h = Q_h + F_{MHA}((Q_h + P_h^2)W_h^q, F_s W_h^k, F_s W_h^v), \label{eq:human_relation}
\end{equation}
\vspace{-25pt}
\begin{align}
&R^o_h = FFN(Q_o),\label{eq:obj_relation_2}   \\
&R^h_o = FFN(Q_h), \label{eq:human_relation_2}  
\end{align} 
where \(R^o_h \in \mathbb{R}^{N_o\times C_d}\) represents the human feature interacting with the object query, and \(R^h_o \in \mathbb{R}^{N_h\times C_d}\) the object feature interacting with the human query, $R \in \mathbb{R}^{N_f \times C_d}$ is defined as $\text{cat}(R_o^h, R_h^o)$. $N_f$ is the sum of $N_o$ and $N_h$.

\subsection{Multi-Head HOI Prediction}\label{subs/decoder_outputs}
\begin{figure}
	\centering
	\includegraphics[width=.45\textwidth]{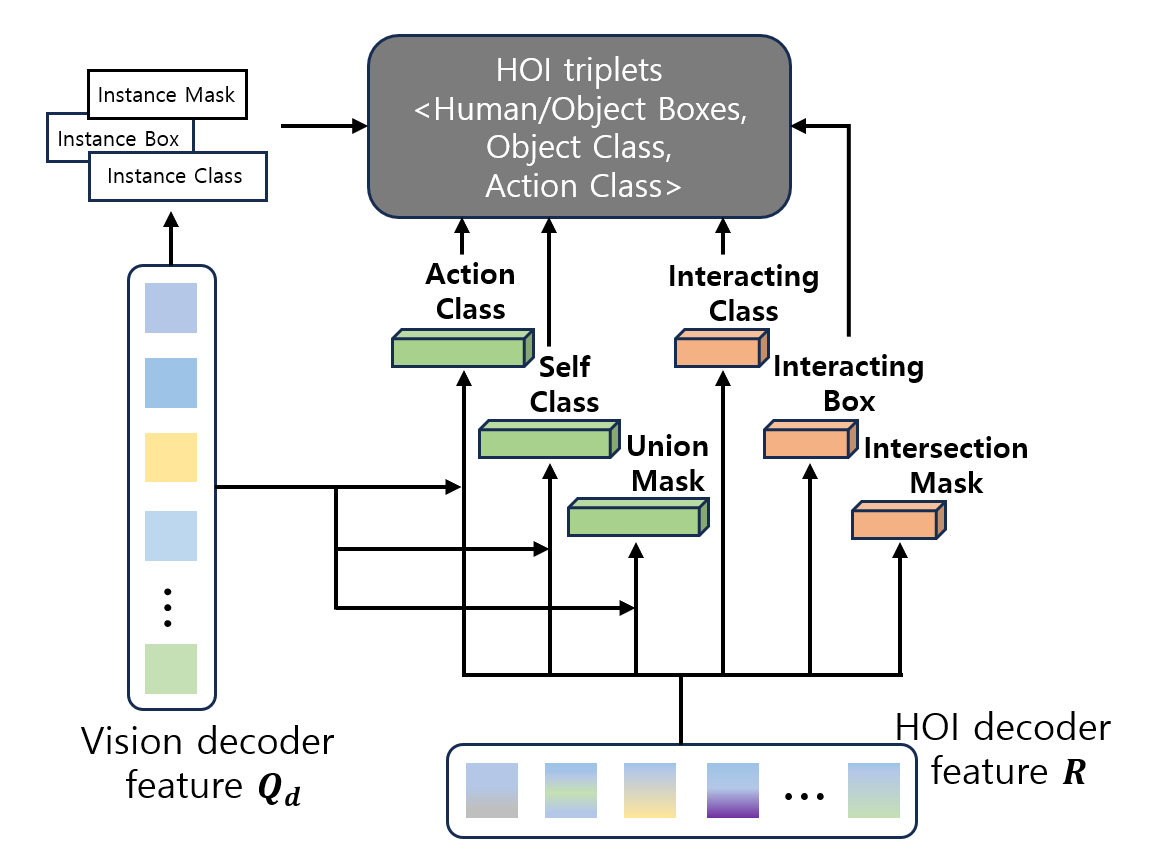}
	\caption{{HOI decoder output heads.}}
	\label{figs/output_head}
    \vspace{-5pt}
\end{figure}

In the conventional two-stage method that creates explicit pairs, there was only a head to determine the action class $c_v$ of the pairwise-query. However, since Seg2HOI does not create explicit pairs, it requires an additional head for the object and human queries to recognize which human and object they are interacting with. Additionally, our framework can perform HOI segmentation. All heads have a relation feature $R$, which is illustrated in Fig.~\ref{figs/output_head}. 
Our process includes a total of six heads: action class $c_v$, self-class $c^{self}_o$, interacting-class $c^{inter}_o$, interacting bounding box $b^{inter}_{\langle h,o \rangle}$, union mask $m^U$, and intersection mask $m^I$ heads for the queries that are interacting. This formulation is as follows:
\begin{align}
c_v &= F_{linear}(\text{cat}(R,Q_d)) \in \mathbb{R}^{N_f \times N_{verb}} \label{eqs:verb_class_for_mask_dino}
,\\
c^{self}_o &= F_{linear}(\text{cat}(R^o_h,Q_d)) \in \mathbb{R}^{N_o \times N_{obj}} \label{eqs:object_class_for_mask_dino}
, \\
c^{inter}_o &= F_{linear}(R^h_o) \in \mathbb{R}^{N_h \times N_{obj}}
,\\
b^{inter}_{\langle h,o \rangle} &= F_{mlp}(R) \in \mathbb{R}^{N_f \times 4}
,\\
m^U &= F_{mlp}(\text{cat}(R,Q_d)) \otimes f_{seg} \in \mathbb{R}^{N_f \times{H\over 4} \times {W\over 4}}  \label{eqs:union_mask}
,\\
m^I &= F_{mlp}(R + Q_d) \otimes f_{seg} \in \mathbb{R}^{N_f \times {H\over 4} \times {W\over 4}} \label{eqs:intersection_mask}
\end{align}
where $F_{linear}$ is the linear layer and $F_{mlp}$ is a feed forward network consist of three linear layers.
\(c^{\text{self}}_o\) is the output from the object branch, while \(c^{\text{inter}}_o\) comes from the human branch. The object branch predicts the self class, thereby obtaining the object class, and the human branch predicts the interacting class, thus obtaining the object class.
The term $b^{inter}_{\langle h,o \rangle}$ predicts interactions similarly to $c^{inter}$. Both the human and object branches are computed in the same way, with $b^{inter}$ encompassing both the human and object boxes, as indicated by the notation ${\langle h,o \rangle}$.
When calculating the HOI mask in Eq.~(\ref{eqs:union_mask}) and (\ref{eqs:intersection_mask}), $f_{seg}$ is multiplied as in Eq.~(\ref{eq/instance_mask}), where the first term is the HOI mask embedding.

\subsection{HOI Mask Pseudo Labeling}\label{subs/pseudo_label}

\begin{figure}
	\centering
	\includegraphics[width=.45\textwidth]{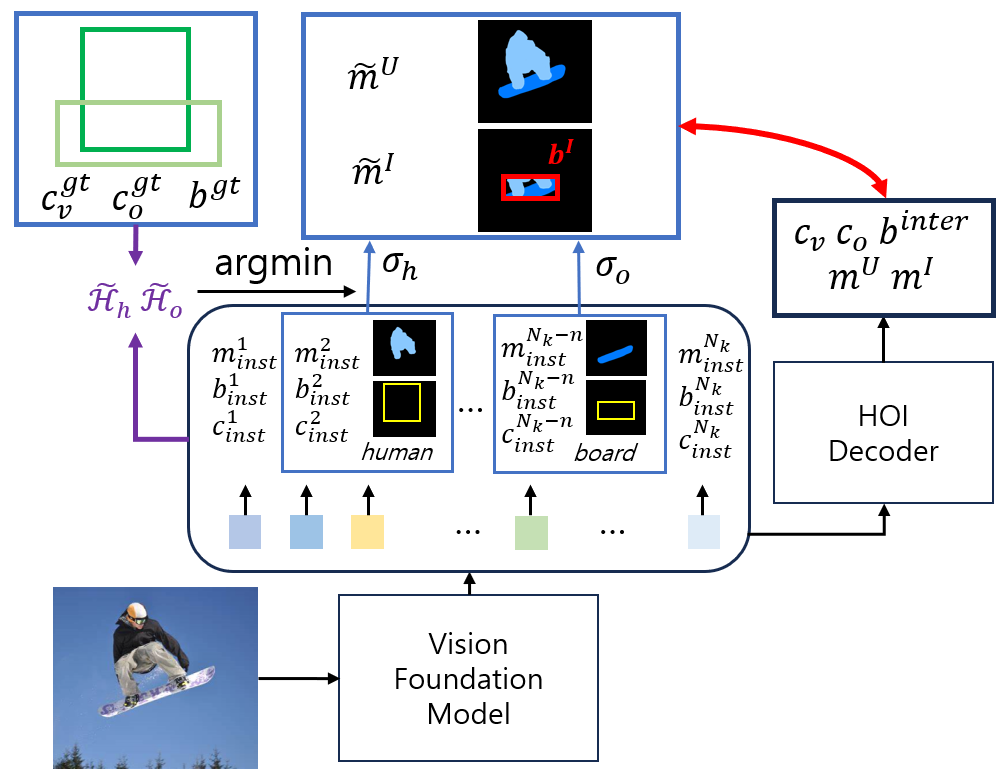}
	\caption{{HOI mask pseudo labeling process }}
	\label{fig/pseudo_label}
\end{figure}

The existing HOI dataset contains bounding box data and class information for humans and objects for the detection task. 
Due to the absence of segmentation masks, it is challenging to train the segmentation-related parts. 
To overcome this issue, we create HOI masks by pseudo-labeling HOI union and intersection masks, leveraging the outputs of a segmentation foundation model. 
We can calculate the loss between the pseudo-labeled HOI mask obtained in this way and the predicted HOI mask from the HOI decoder. 
We generate pseudo labels by comparing the instance mask with the ground truth bounding box, with further details provided below.
Let \( b^{gt}_h \) and \( b^{gt}_o \) be the ground truth bounding boxes of the human and object pair, respectively. Define \( m^{pred}_h \), \( m^{pred}_o \subset m_{inst} \) as the predicted human and object instance masks, respectively. Let \( b^m_h \) and \( b^m_o \) be the bounding boxes obtained from the min-max indexes of these masks. We can then obtain the HOI mask through the following process. This process is illustrated in Fig.~\ref{fig/pseudo_label}.
The costs for the human and object are denoted as 
\begin{align}
\widetilde{\mathcal{H}}_h = \beta_b\mathcal{H}^{(b)}(b^{gt}_h, b_h^m) + \beta_u\mathcal{H}^{(u)}(b^{gt}_h, b_h^m)
, \\
\widetilde{\mathcal{H}}_o = \beta_b\mathcal{H}^{(b)}(b^{gt}_o, b_o^m) + \beta_u\mathcal{H}^{(u)}(b^{gt}_o, b_o^m)  ,
\end{align}
where \(\beta_b\) and \(\beta_u\) are hyperparameters.

To obtain the masks for each instance of human and object from the output of the segmentation model, we first need to find the query index with the minimum cost value for each instance from the foundation model's output, as 
\begin{equation}
\sigma_h = \arg\min \widetilde{\mathcal{H}}_h, \ \sigma_o = \arg\min \widetilde{\mathcal{H}}_o.
\label{eq:instance_index}
\end{equation}
Based on the query indices, we can obtain the union mask and intersection mask.
\begin{align}
\widetilde{m}^U &= m^{pred}_h(\sigma_h) \cup  m^{pred}_o(\sigma_o) \label{eq:hoi_union_mask}
, \\
\widetilde{m}^I &= \text{crop}(\widetilde{m}^U, b^{I}), 
\label{eq:hoi_intersection_mask}\\
b^{I} &= (b^m_h[\sigma_h] \pm \gamma) \cap  (b^m_o[\sigma_o] \pm \gamma),
\end{align}
where the union mask \(\widetilde{m}^U\) is a union of \(m^{\text{pred}}_h\) and \(m^{\text{pred}}_o\). The intersection mask \(\widetilde{m}^I\) is the cropped region of the union mask by the intersection area \(b^{I}\) which is the expanded bounding box \([x_{\min}, y_{\min}, x_{\max}, y_{\max}]\) with a value of \(\gamma\), as depicted in Fig. \ref{fig/pseudo_label}.
The union and intersection masks are used for Hungarian algorithm in HOI decoder training and for HOI segmentation mask training.

\subsection{Training and Inference}\label{subs/training}
Since we utilize a pretrained vision foundation model in a frozen state, there is no need for additional training to obtain instance bounding boxes and class labels. Instead, it is sufficient to train only the HOI decoder. Nonetheless, the proposed methods effectively transfer relevant information from the vision foundation model to the HOI decoder. 
It is worth noting that all the proposed components together create a key distinction of the Seg2HOI framework compared to other HOI methods: it outputs not only the set of detected HOIs but also the segmentation masks for the human-object pairs involved in the interactions. 
As demonstrated in our experiments, the high-quality HOI quadruplets $Z = \{m^{s}, b_{h}, (b_{o}, c_{o}), v\}$ can be obtained efficiently.
These HOI quadruplets have broad applicability, enhancing functionalities in areas such as scene understanding, robotics, and augmented reality.

As shown in Sec.~\ref{subs/decoder_outputs} and Table \ref{fig:query_input_features}, \(c^{\text{self}}_o\) and \(c^{\text{inter}}_o\) both predict the object class, but since each of them has a different dimensions compared to the others, we concatenate \(c^{\text{self}}_o\) and \(c^{\text{inter}}_o\) to form \(c_o\) for more efficient training.
Additionally, to enhance performance and support applications, the HOI decoder applies auxiliary loss functions to facilitate the learning of detailed pixel information through HOI segmentation, such as union masks and intersection masks.
The proposed training method follows the approach of previous transformer-based methods \citep{tamura2021qpic, zhang2021mining, kim2023relational}. We adopt Hungarian Matching to assign ground truth HOI triplets to predictions, and we extend the formulation by adding an HOI segmentation term, resulting in the following final equation:
\begin{equation}
\begin{aligned}
&\mathcal{H}_{\text{match}}^{seg2hoi} = \ \alpha_v\mathcal{H}^{(a)}(c_v^{gt}, c_v) + \alpha_c\mathcal{H}^{(c)}(c_o^{gt}, c_o)  \\
&+ \alpha_b\mathcal{H}^{(b)}(b^{gt}, b^{inter}_{\langle h,o \rangle}) + \alpha_u\mathcal{H}^{(u)}(b^{gt}, b^{inter}_{\langle h,o \rangle})  \\
&+ \sum_{k \in \{U, I\}} \left( \alpha_{e}\mathcal{H}^{(e)}(\widetilde{m}^k, m^k) + \alpha_d\mathcal{H}^{(d)}(\widetilde{m}^k, m^k) \right),
\end{aligned}
\end{equation} 
where \(\mathcal{H}^{(a)}\) is the cost related to the interacting action class, and \(\mathcal{H}^{(b)}\), \(\mathcal{H}^{(u)}\) are cost functions as defined in Eq.~(\ref{eq/fundamental_task_match_cost_function}). 
The detailed equations can be found in the \citep{tamura2021qpic}.  \(\mathcal{H}^{(e)}\), and \(\mathcal{H}^{(d)}\) are cost functions related with masks as defined in Eq.~(\ref{eq/fundamental_task_match_cost_function}). Regarding the mask-related costs, it can be observed that we deal with the HOI union mask, not instance mask. Optionally, the cost of the intersection mask can be added to cost function $\mathcal{H}_{\text{match}}^{seg2hoi}$.

\begin{table}[t]
    \centering
    \caption{Head input table. The $*$ denotes an auxiliary term for training or application purposes, which is not included in the HOI triplet.}
    \begin{tabular}{lccc}
        \toprule
        Head & & Object query & Human query \\
        \midrule
        Action Class & $c_v$ &  Cat($R^o_h$, $Q_d$) & Cat($R^h_o$, $Q_d$) \\
        Self Class & $c^{self}_o$ & Cat($R^o_h$, $Q_d$) & - \\
        Interacting Class & $c^{inter}_o$ & - & $R^h_o$ \\
        Interacting Box & $b^{inter}_{\langle h,o \rangle}$ & $R^o_h$ & $R^h_o$ \\
        Union Mask & $m^U$ & * Cat($R^o_h$, $Q_d$) & * Cat($R^h_o$, $Q_d$) \\
        Intersection Mask & $m^I$ & * $R^o_h$ + $Q_d$ &* Cat($R^h_o$, $Q_d$) \\
        \bottomrule
    \end{tabular} 
    \label{fig:query_input_features}
\end{table}

The training loss directly guides the HOI segmentation mask instead of the instance mask, enabling the HOI decoder to learn more effectively, resulting in the following formulation.
\begin{equation}
\begin{aligned}
&\mathcal{L}_{\text{loss}}^{seg2hoi} = \ \lambda_v \mathcal{L}^{(a)}(c_v^{gt}, c_v) + \lambda_c\mathcal{L}^{(c)}(c_o^{gt}, c_o)  \\
&+ \lambda_b\mathcal{L}^{(b)}(b^{gt}, b^{inter}_{\langle h,o \rangle}) + \lambda_u\mathcal{L}^{(u)}(b^{gt}, b^{inter}_{\langle h,o \rangle})  \\
&+ \sum_{k \in \{U, I\}} \left( \lambda_{e}\mathcal{L}^{(e)}(\widetilde{m}^k, m^k) + \lambda_d\mathcal{L}^{(d)}(\widetilde{m}^k, m^k) \right),
\end{aligned}
\end{equation}
where all parts except for the mask follow the loss function of \citep{tamura2021qpic}, while the mask related loss incorporates point loss for efficiency, as done in \citep{li2023mask, cheng2022masked, cheng2022pointly}.

During inference, we set the prediction results of HOI instances \( \langle b_h^i, b_o^i, \text{argmax}_k\ c_o^i(k), c_v^i(j) \rangle\) from each \(i\)-th query and \(j\)-th action.  We define the HOI confidence score $c^i_{hoi}$ as \(\{\max_k c_o^i(k)\}^{\lambda} \sigma(c_v^i(j))\), where \(\lambda \in [0,1]\) is a hyperparameter.

\begin{figure}
	\centering
	\includegraphics[width=.45\textwidth]{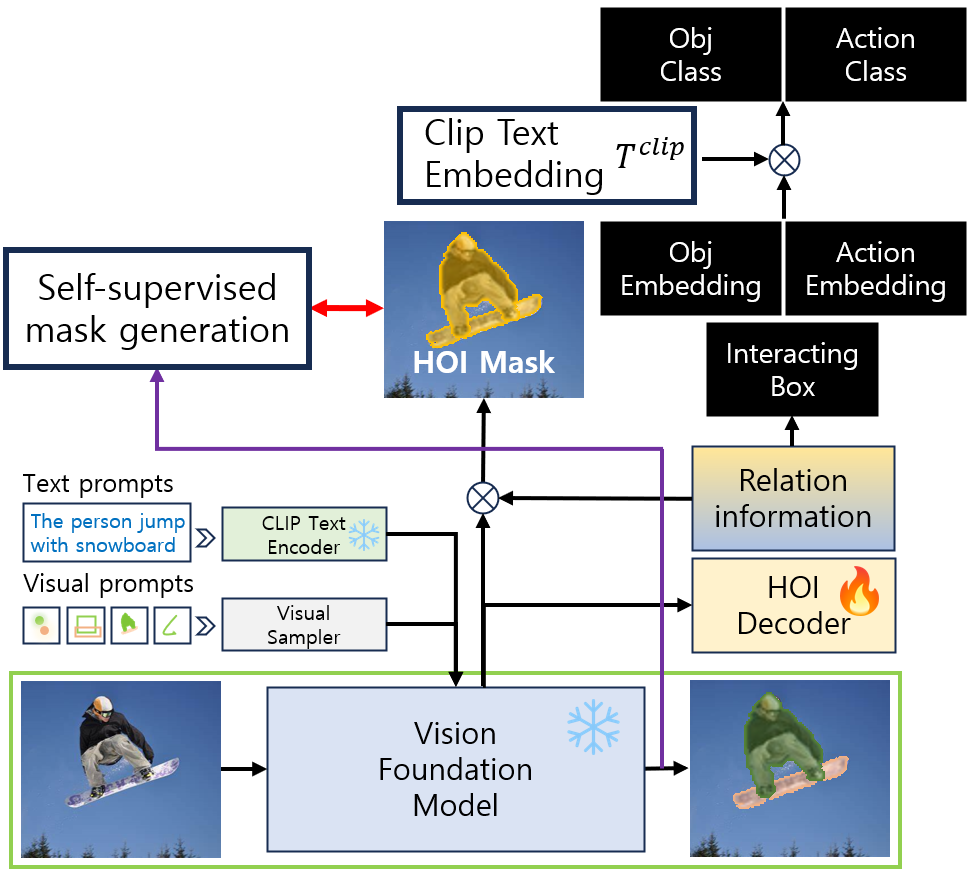}
	\caption{{Illustration of our Seg2HOI framework architecture for HOI recognition,
    which consists of a segmentation foundation model and an HOI decoder, providing a methodology to obtain relation results.}}
	\label{figs:seem2hoi}
\end{figure}

\section{Interactive Quadruplet Image Segmentation Extension}\label{sec:seem2hoi}
In this section, we demonstrate that by integrating the proposed methodology with an open-vocabulary capable vision language foundation model, the resulting open-vocabulary HOI detection and segmentation can operate interactively even with diverse prompts. Thus, we utilize CLIP \citep{radford2021learning} to embed object classes and interacting action classes using the CLIP text encoder for training. This framework is illustrated in Fig.~\ref{figs:seem2hoi}. 

\textbf{Text Embedding for Classifier.} To train the classifier using CLIP text embeddings, we need to obtain text embeddings for the object class label $c_o$ and the interacting action (verb) class $v$. First, as done in \citep{Liao_2022_CVPR, ning2023hoiclip, ye2023diffusion}, we convert the class label and interacting action class label into sentences using hand-crafted templates, $i.e.$, "A photo of a person and a/an [Object]" for the object classifier, "A photo of a person [Verb-ing]". These sentences are then passed through the CLIP text encoder to obtain the object category embedding \(T_o^{clip} \in \mathbb{R}^{N_{obj} \times C_d}\) and the interacting action category embedding \(T_v^{clip} \in \mathbb{R}^{N_{verb} \times C_d}\). 
These obtained category embeddings can be used for classification. However, unlike the direct classification through \(F_{linear}\) in Eq (\ref{eqs:verb_class_for_mask_dino}) and  (\ref{eqs:object_class_for_mask_dino}), this process performs classification by calculating similarity scores between embeddings derived from visual features and the text embeddings. This process is as follows:
\begin{align}
E_v &= F_{linear}(\text{cat}(R,Q_d)) \in \mathbb{R}^{N_f \times C_d},  \\
E^{self}_o &= F_{linear}(\text{cat}(R^o_h,Q_d)) \in \mathbb{R}^{N_o \times C_d}, \\
E^{inter}_o &= F_{linear}(R^h_o) \in \mathbb{R}^{N_h \times C_d},  \\
E_o &= \text{cat}(E^{self}_o, E^{inter}_o) \in \mathbb{R}^{N_f \times C_d} 
\end{align} 
where $E_v, E_o$ are action embeddings and object embeddings derived from visual features, repectively. 
\begin{align}
c_v &= \text{sim}\left(E_v, T_v^{clip}\right) \in \mathbb{R}^{N_f \times N_{verb}},  \\
c_o &= \text{sim}\left(E_o, T_o^{clip}\right) \in \mathbb{R}^{N_f \times N_{obj}} 
\end{align} 
where \text{sim} represents the cosine similarity function, i.e., $\text{sim}\left(E, T^{clip}\right) = E \cdot {T^{clip}}^\top / ({\|E\|\|T^{clip}\|})$.
For the remaining heads that do not utilize text embedding, such as the HOI mask and interacting box, these can be obtained in the same manner as described in Sec.~\ref{subs/decoder_outputs}.

\textbf{Prompt-Based Image Segmentation.}
The text and visual prompts, obtained from the CLIP text encoder and visual sampler \citep{zou2024segment}, respectively, generate prompt queries within the same visual embedding space. These prompts are shown in Fig.~\ref{fig:seem_qualitative_visual} and \ref{fig:seem_qualitative_text}, where the visual and text prompts are illustrated, respectively. Leveraging the strong alignment of vision foundation model, these queries are then passed through our HOI decoder and converted into HOI relation features without additional prompt-specific training.

During inference, samples are generated based on the similarity between the features of these prompts and the features of the HOI decoder, as follows: 
\begin{equation}
\sigma^{visual} = \arg\max \text{sim}\left(E^U, T^{visual}\right), \label{eq:visual_prompt}
\end{equation}
\vspace{-22.5pt}
\begin{equation}
\sigma^{text} = \arg\max \left(\text{sim}\left(E_o, T^{text}\right) * \text{sim}\left(E_v, T^{text}\right)\right), \label{eq:text_prompt}
\end{equation}
where $E^U$ is the first term in Eq.~(\ref{eqs:union_mask}), $T^{visual}$ and $T^{text}$ are the features of the visual and text prompts, respectivly, obtained after passing through the HOI decoder. 
$\sigma^{visual}$ and $\sigma^{text}$ are the indices of the maximum similarity between the prompt and the HOI decoder features, and these values are smaller than $N_f$.

Through this process, the corresponding quadruplet for the given query index can be obtained from various prompts. 
While this can be implemented in other ways, this paper demonstrates it using Eq.~(\ref{eq:visual_prompt}) and Eq.~(\ref{eq:text_prompt}).

\section{Experiments}
Our main objective is to propose a quadruplet method; however, since the triplet is a subset of the quadruplet, we quantitatively compare the performance of existing HOI methods proposed for the triplet with our approach in this section. Qualitatively, we visualize the quadruplet to provide further insights.
\subsection{Experimental Settings}
\textbf{Datasets.} We evaluate our model on two public benchmark, HICO-DET \citep{chao2018learning} and V-COCO \citep{gupta2015visual}. HICO-DET offers 47,776 images with 151,276 HOI instances (38,118 images with 117,871 annotated instances of human-object pairs for training and 9,658 images with 33,405 annotated instances of human-object pair for testing). It includes 80 object classes, identical to in the Microsoft Common Objects in Context (MS-COCO) \citep{lin2014microsoft} dataset, 117 action classes, and 600 HOI classes in the datasets that group into three categories: Full (all HOI classes), Rare (138 classes with fewer than ten instances) and Non-Rare (462 classes with more than ten instances). V-COCO provides 10,346 images with 16,199 person instances (2,533 images for training, 2,867 for validation, and 4,946  for testing). It contains 80 object classes and 29 action categories and is a subset of MS-COCO dataset. The objects are divided into two types: "object" and "instrument".

\textbf{Metrics.} We report mean average precision (mAP) for both datasets, following the settings in \citep{chao2018learning}. An HOI set prediction is considered a true positive if the predicted human and object boxes both have an IoU greater than 0.5 with the ground truth, and both the object class and action class are correct. For HICO-DET, following \citep{chao2018learning}, we report mAP for two types: Default and Known Object. In the former case, average precision (AP) is calculated using all test images, whereas in the latter case, it is calculated only using images that contain the object class corresponding to each AP. Additionally, we report mAP for the three previously mentioned groups (Full, Rare, Non-Rare). In V-COCO, we report role mAPs for two scenarios \citep{gupta2015visual}: scenario 1 (S1), which includes the object, and scenario 2 (S2), which ignores the predictions of an object bounding box in these cases.

\textbf{Zero-Shot Evaluation.} We evaluate the Seg2HOI framework on HOI zero-shot detection in HICO-DET, as it can be applied to various segmentation foundation models and can naturally extend to models capable of open-vocabulary learning. Following \citep{hou2020visual, hou2021affordance, liao2022gen}, we conduct experiments on rare first unseen composition (RF-UC), non-rare first UC (NF-UC), and unseen object (UO). 
Specifically, in the UC setting, we exclude certain HOI triplets; in the UO setting, specific object classes; and in the UV setting, specific action classes from the training set. We evaluate 120 unseen, 480 seen, and all 600 categories for the UC setting. The Rare First approach prioritizes categories from tail HOIs, while the Non-Rare First approach prioritizes head categories.
For the UO setting, we evaluate 100 unseen and 500 seen HOI categories, including 12 unseen objects among the 80 object classes, following the approach in \citep{hou2020visual}. 
Additionally, we verify that our model accurately detects HOI quadruplets for zero-shot compositions from various text prompts using the model trained on a zero-shot HOI dataset, as detailed in Sec.~\ref{quadruplet_interactive_hoi_segmentation_experiments}.

\begin{table}[t]
    \centering
    \caption{Comparison of performance on V-COCO dataset. $\dagger$ indicates results obtained using both the HOI union mask and intersection mask for training. Results without this mark report the use of the union mask only. For convenience, Mask-DINO is referred to as M-DINO.}
    \begin{tabular}{lccc}
        \toprule
        \multirow{2}{*}{Method} & \multirow{2}{*}{Backbone} & \multicolumn{2}{c}{V-COCO}  \\
        \cmidrule(lr){3-4}
         & & $\text{AP}^{S1}_{\text{role}}$ & $\text{AP}^{S2}_{\text{role}}$ \\
        \midrule
        \multicolumn{4}{l}{\textit{One-Stage Methods}} \\
        InteractNet \citep{gkioxari2018detecting} & ResNet-50-FPN & 40.0 & - \\
        iCAN \citep{gao2018ican}& ResNet-50 & 45.3 & 52.4 \\
        HOTR \citep{kim2021hotr} & ResNet-50 & 55.2 & 64.4 \\
        QPIC \citep{tamura2021qpic} & ResNet-101 & 58.8 & 61.0 \\
        CDN \citep{zhang2021mining}& ResNet-101 & 63.9 & 65.9 \\
        QAHOI \citep{chen2023qahoi}& ResNet-50 & 58.2 & 58.7 \\
        FGAHOI \citep{ma2023fgahoi}& ResNet-50 & 59.0 & 59.3 \\
        FGAHOI \citep{ma2023fgahoi}& Swin-T & 60.5 & 61.2 \\
        GEN-VLKT \citep{Liao_2022_CVPR}& ResNet-50 & 62.4 & 64.5 \\
        MUREN \citep{kim2023relational}& ResNet-50 & \textbf{68.8} & \textbf{71.0} \\ 
        \midrule
        \multicolumn{4}{l}{\textit{Two-Stage Methods}} \\
        VCL \citep{hou2020visual}& ResNet-50 & 48.3 & - \\
        IDN \citep{li2020hoi}& ResNet-50 & 53.3 & 60.3 \\
        UPT \citep{zhang2022efficient} & ResNet-101-DC5 & 61.3 & 67.1 \\
        ViPLO \citep{park2023viplo}& ViT-B/32 & 60.9 & 66.6 \\ 
        STIP \citep{zhang2022exploring} & ResNet-50 & 66.0 & 70.7 \\
        PViC \citep{zhang2023exploring} w/ DETR  & ResNet-50 & 62.8 & 67.8 \\
        PViC \citep{zhang2023exploring} w/ H-DETR & Swin-L & 64.1 & 70.2 \\
        \rowcolor{gray!20}Seg2HOI w/ M-DINO  $\dagger$ & ResNet-50 & \textbf{67.3} & \textbf{69.2}\\
        \rowcolor{gray!20}Seg2HOI w/ M-DINO & Swin-L & \textbf{70.7} & \textbf{73.0} \\
        \rowcolor{gray!20}Seg2HOI w/ SEEM & Focal-T & 58.1 & 59.9 \\
        \rowcolor{gray!20}Seg2HOI w/ SEEM & Focal-L & 66.3 & 68.4 \\
        \bottomrule
    \end{tabular} 
    \label{table:VCOCO_eval}
    \vspace{-5pt}
\end{table}
\textbf{Implementation Details.} 
To demonstrate effectiveness of the Seg2HOI framework, we choose the Mask-DINO \citep{li2023mask} as a frozen vision foundation model and the SEEM \citep{zou2024segment} as a frozen vision language foundation model, respectively. 
During training, to ensure the detection of interactions between a single human and multiple objects, we repeated the human query four times when constructing the human decoder’s query.
Consequently, the maximum number of human queries is set to 60, and if the number of humans is insufficient, attention is implemented with masking. For the focal loss, we set $\alpha = 0.5$ and $\gamma = 2.0$. 
The hyperparameter $\lambda$ in the inference process is set to $0.5$. 
The HOI decoder follows the hidden dimension of the pretrained segmentation model and is composed of 6 stacked layers. 
The entire architecture is trained on four NVIDIA A6000 Ada GPUs using the AdamW optimizer with a weight decay of $10^{-4}$. 
The mini-batch size is set to 16 for HICO-DET and 8 for V-COCO. 
We only trained the HOI decoder, initializing the learning rate to $1 \times 10^{-4}$. 
The maximum number of training epochs is 70, with learning rate reductions at the 50th and 60th epochs, reduced by a factor of 5 each time. 
The hyperparameters for the Hungarian costs, $\lambda_v, \lambda_c, \lambda_b, \lambda_u, \lambda_U, \lambda_I$, are set to $5.0$, $4.0$, $5.0$, $2.0$, $2.0$, and $0.1$, respectively. 
Similarly, the hyperparameters for the loss weights, $\alpha_v, \alpha_c, \alpha_b, \alpha_u, \alpha_U, \alpha_I$, are set to $5.0$, $5.0$, $3.0$, $4.0$, $2.0$, and $0.1$, respectively.

\begin{table*}[t]
    \centering
    \caption{Comparison of different methods on HICO-DET dataset. $^{*}$ means the results of experiments conducted directly for the purpose of comparing with other studies.}
    \resizebox{\columnwidth}{!}{
    \begin{tabular}{lcccccccc}
        \toprule
        \multirow{2}{*}{Method} & \multirow{2}{*}{Detector} & \multirow{2}{*}{Backbone} & \multicolumn{3}{c}{Default Setting} & \multicolumn{3}{c}{Known Objects Setting} \\
        \cmidrule(lr){4-6} \cmidrule(lr){7-9}
        & & & Full & Rare & Non-rare & Full & Rare & Non-rare \\
        \midrule
        PPDM \citep{liao2020ppdm} & HICO-DET & Hourglass-104 & 21.94 & 13.97 & 24.32 & 24.81 & 17.09 & 27.12 \\
        VCL \citep{hou2020visual}& HICO-DET & ResNet-50 & 23.63 & 17.21 & 25.55 & 25.89 & 19.12 & 28.03 \\
        ATL \citep{hou2021affordance}& HICO-DET & ResNet-50 & 23.81 & 17.43 & 25.72 & 27.38 & 22.09 & 28.96 \\
        HOTR \citep{kim2021hotr}& HICO-DET & ResNet-50 & 25.10 & 17.34 & 27.42 & - & - & - \\
        QPIC \citep{tamura2021qpic} & HICO-DET & ResNet-101 & 29.90 & 23.92 & 31.69 & 32.38 & 26.06 & 34.27 \\
        CDN \citep{zhang2021mining} & HICO-DET & ResNet-101 & 32.07 & 27.19 & 33.53 & 34.79 & 29.48 & 36.38 \\
        UPT \citep{zhang2022efficient} & HICO-DET & ResNet-101-DC5 & 32.62 & 28.68 & 33.71 & 36.08 & 31.47 & 37.47 \\
        QAHOI \citep{chen2023qahoi}& HICO-DET & Swin-T & 28.47 & 22.44 & 30.27 & 30.99 & 24.83 & 32.84  \\
        QAHOI \citep{chen2023qahoi}& HICO-DET & Swin-L & 35.78 & 29.80 & 37.56 & 37.59 & 31.66 & 39.36  \\
        FGAHOI \citep{ma2023fgahoi} & HICO-DET & Swin-T & 29.94 & 22.24 & 32.34 & 32.48 & 24.16 &34.97 \\
        FGAHOI \citep{ma2023fgahoi}& HICO-DET & Swin-L & 37.18 & 30.71 & 39.11 &38.93 &31.93 &41.02 \\
        GEN-VLKT \citep{Liao_2022_CVPR}& HICO-DET & ResNet-50 & 33.75 & 29.25 & 35.10 & 36.78 & 32.75 & 37.99 \\
        MUREN \citep{kim2023relational}& HICO-DET & ResNet-50 & 32.87 & 28.94 & 34.12 & 35.52 & 30.88 & 36.91 \\
        STIP \citep{zhang2022exploring} w/ DETR & HICO-DET & ResNet-50 & 32.22 &28.15& 33.43 & 35.29 &31.43 & 36.45 \\
        PViC \citep{zhang2023exploring} w/ DETR & HICO-DET & ResNet-50 & 34.69 &32.14& 35.45 & 38.14 &35.38 & 38.97 \\
        PViC \citep{zhang2023exploring} w/ H-DETR & HICO-DET & Swin-L & \textbf{44.32} & \textbf{44.61} & \textbf{44.24} & \textbf{47.81} & \textbf{48.38} & \textbf{47.64}  \\
        \midrule
        InteractNet \citep{gkioxari2018detecting} & COCO & ResNet-50-FPN & 9.94 & 7.16 & 10.77 & - & - & - \\
        iCAN \citep{gao2018ican} & COCO & ResNet-50 & 14.84 & 10.45 & 16.15 & 16.26 & 11.33 & 17.73 \\
        VCL \citep{hou2020visual} & COCO & ResNet-50 & 19.43 & 16.55 & 20.29 & 22.00 & 19.09 & 22.87 \\
        VSGNet \citep{ulutan2020vsgnet}& COCO & ResNet-152 & 19.80 & 16.05 & 20.91 & - & - & - \\
        PastaNet \citep{li2020pastanet} & COCO & ResNet-50 & 22.65 & 21.17 & 23.09 & 24.53 & 23.00 & 24.99 \\
        HOTR \citep{kim2021hotr}& COCO & ResNet-50 & 23.46 & 16.21 & 25.62 & - & - & - \\
        STIP \citep{zhang2022exploring}w/ DETR & COCO & ResNet-50 & 28.81 &27.55& 29.18 & 32.28 &31.07 & 32.64 \\
        PViC \citep{zhang2023exploring} w/ $\text{DETR}^{*}$ & COCO & ResNet-50 & 29.19 & \textbf{28.53} & 29.38 & \textbf{33.08} & \textbf{32.12} & 33.36 \\
        \rowcolor{gray!20} Seg2HOI w/ MaskDINO & COCO & ResNet-50 & \textbf{29.67} &24.38 &\textbf{31.24}& 32.61 & 27.33 & \textbf{34.18} \\
        \rowcolor{gray!20} Seg2HOI w/ MaskDINO & COCO & Swin-L & \textbf{37.98}  & \textbf{35.58} & \textbf{38.69} & \textbf{40.24} & \textbf{38.06} & \textbf{40.89} \\
        \rowcolor{gray!20} Seg2HOI w/ SEEM & COCO & Focal-T & 26.78 & 22.71 & 27.99 & 29.14 & 25.17 & 30.32 \\
        \rowcolor{gray!20} Seg2HOI w/ SEEM & COCO & Focal-L & 35.70  & 35.07 & 35.89 & 37.91 & 37.32 & 38.08 \\
        \bottomrule
    \end{tabular} }
    \label{table:HICO_eval}
    \vspace{-10pt}
\end{table*}

\subsection{Closed Vocabulary Results}
This subsection presents experiments performed within the closed vocabulary setting. The results, assessed using the official evaluation code and benchmarked against state-of-the-art methods~\citep{Liao_2022_CVPR, NEURIPS2021_8f1d4362, zhang2023exploring, tamura2021qpic,  kim2023relational}, are shown in Table~\ref{table:VCOCO_eval} and Table~\ref{table:HICO_eval}, respectively.

\textbf{Comparison with State-of-the-Arts HOI.} 
As shown in Table~\ref{table:VCOCO_eval}, for the V-COCO dataset, Seg2HOI achieves state-of-the-art performance among previous two-stage HOI models, with a role AP of 67.3 in Scenario 1 and 69.2 in Scenario 2.
As shown in Table~\ref{table:HICO_eval}, for HICO-DET, Seg2HOI shows the best performance when the detector is fine-tuned only on the MS-COCO dataset. 
PViC is a state-of-the-art model among two-stage methods;
however, the results reported in \citep{zhang2023exploring} were obtained using a 2-phase training approach: The backbone was fine-tuned on HICO-DET as a detector, then frozen before training the HOI component. 
We point out that recent HOI papers are primarily focused on improving performance by fine-tuning the backbone on HICO-DET, which we find to be a significant concern. 
This approach modifies the weights of the foundation model, potentially compromising the performance on the original fundamental task. 
Therefore, we have added experiments with PViC using the pretrained weights to preserve the original performance of the detector. 
Based on the official PViC code, when comparing the PViC model trained on the COCO dataset with Seg2HOI, we obtain highly comparable results. 
Notably, in the default setting for non-rare categories, Seg2HOI achieves a significant mAP gain of 6.3$\%$ over PViC. 
Furthermore, Seg2HOI outperforms the remaining COCO-based HOI models by a large margin. 
These experimental results demonstrate that Seg2HOI delivers state-of-the-art or reasonably similar performance in HOI detection, while crucially maintaining the performance of other important tasks such as segmentation and object detection. This highlights Seg2HOI's advantage in offering a balanced approach, unlike other models designed exclusively for HOI detection.

\begin{figure*}
    \centering
    \includegraphics[width=1\textwidth]{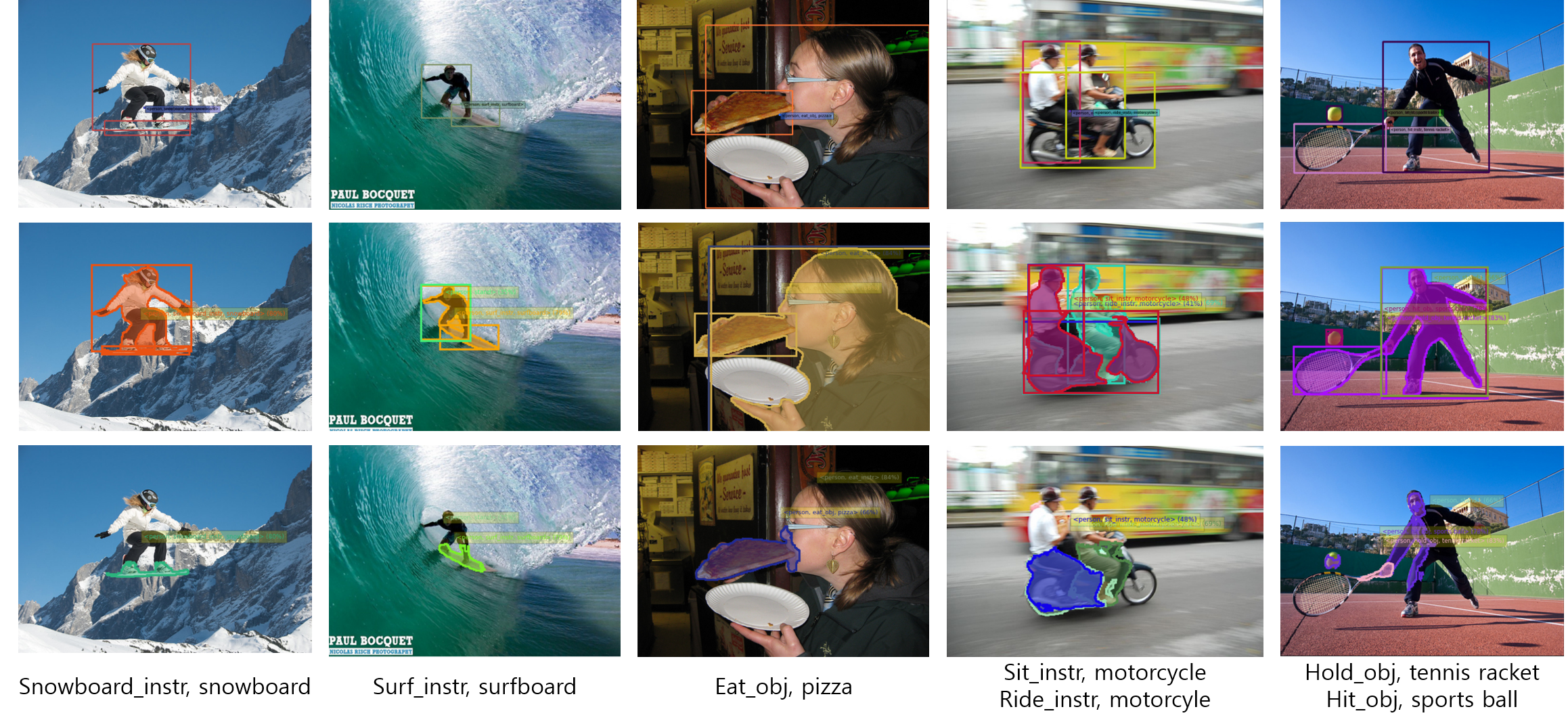}
    \vspace{-15pt}
    \caption{We showcase qualitative results proposed quadruplets. The first row represents the ground truth triplet data, the second row shows the Seg2HOI quadruplet output with HOI union mask, and the third row displays the HOI intersection mask.}
    \label{fig:seg2hoi_qualitative_result}
    \vspace{-10pt}
\end{figure*}

\textbf{Quadruplet Qualitative Results.} We showcase the results of the proposed method. As shown in Fig.~\ref{fig:seg2hoi_qualitative_result}, the method performs well on a wide range of HOI compositions, demonstrating successful HOI segmentation.    
In Fig.~\ref{fig:seg2hoi_qualitative_result}, the first row represents the triplet, while our quadruplet is shown in the second row, demonstrating a clean output that includes HOI union segmentation. The third row visualizes the HOI intersection mask, demonstrating that our proposed pseudo-labeling method effectively trains the HOI decoder and extracts HOI mask features that align well with the segmentation branch’s features, without finetuning of the foundation model.

\subsection{Open Vocabulary Results}

All experiments were performed with Seg2HOI employing MaskDINO (Resnet-50) and SEEM (focal-tiny), using only the union mask. We train the model without HOI segmentation mask strategy as a baseline.
As shown in Table~\ref{table:HOI_zeroshot_eval}, Seg2HOI showcases comparable results to previous method in zero-shot HOI detection for UC and UO settings.

\begin{table}[ht]
\centering
\caption{Comparison of different methods on various types of unseen/seen HOI categories for \textbf{zero-shot HOI detection}. RF stands for rare first, NF stands for non-rare first, and UC, UO, UV represent unseen composition, unseen object, and unseen verb, respectively. FT refers to fine-tuning the backbone or detector on the HICO-DET dataset. * means only the detected boxes are used without object identity information from the detector.}
\begin{tabular}{lccccc}
\hline
\text{Method} & FT & \text{Type} & \text{Unseen} & \text{Seen} & \text{Full} \\
\hline
Shen \textit{et al.} \citep{maraghi2021scaling} & O & UC & 5.62 & - & 6.26 \\
FG \citep{bansal2020detecting} & X & UC & 10.93 & 12.60 & 12.26 \\
ConsNet \citep{liu2020consnet} & O & UC & 16.99 & 20.51 & 19.81 \\
\hline
VCL \citep{hou2020visual} & X & RF-UC & 7.55 & 18.84 & 16.58 \\
VCL \citep{hou2020visual} & O & RF-UC & 10.06 & 24.28 & 21.43 \\
ATL \citep{hou2021affordance} & O & RF-UC & 9.18 & 24.67 & 21.57 \\
FCL \citep{hou2021detecting} & O & RF-UC & 13.16 & 24.23 & 22.01 \\
$\text{GEN-VLKT}_s$ \citep{hou2021affordance} & O & RF-UC & 21.36 & 32.91 & 30.56 \\
\rowcolor{gray!20} baseline w/SEEM & X & RF-UC & 17.78 & 24.30 & 22.96 \\
\rowcolor{gray!20} Seg2HOI w/SEEM & X & RF-UC & 20.10 & 25.38 & 24.32 \\
\rowcolor{gray!20} Seg2HOI w/M-DINO & X & RF-UC & 16.33 & 27.82 & 25.52 \\
\hline
VCL \citep{hou2020visual}& X & NF-UC & 9.13 & 13.67 & 12.76 \\
VCL \citep{hou2020visual}& O & NF-UC & 16.22 & 18.52 & 18.06 \\
ATL \citep{hou2021affordance}& O & NF-UC & 18.25 & 18.78 & 18.67 \\
FCL \citep{hou2021detecting}& O & NF-UC & 18.66 & 19.55 & 19.37 \\
$\text{GEN-VLKT}_s$ \citep{hou2021affordance} & O & NF-UC & 25.05 & 23.38 & 23.71 \\
\rowcolor{gray!20} baseline w/SEEM & X & NF-UC & 18.83 & 18.16 & 18.30 \\
\rowcolor{gray!20} Seg2HOI w/SEEM & X & NF-UC & 20.03 & 21.33 & 21.07 \\
\rowcolor{gray!20} Seg2HOI w/M-DINO & X & NF-UC & 20.26 & 22.08 & 21.72  \\
\hline
$\text{ATL}^*$ \citep{hou2021affordance}& O & UO & 0.00 & 13.67 & 11.39 \\
$\text{ATL}^*$ \citep{hou2021affordance}& X & UO & 5.05 & 14.69 & 13.08 \\
$\text{FCL}^*$ \citep{hou2021detecting}& O & UO & 0.00 & 13.71 & 11.43 \\
$\text{GEN-VLKT}_s$ \citep{hou2021affordance} & O & UO & 10.51 & 28.92 & 25.63 \\
\rowcolor{gray!20} baseline w/SEEM & X & UO & 4.83 & 14.87 & 13.64 \\
\rowcolor{gray!20} Seg2HOI w/SEEM & X & UO & 9.17 & 21.55 & 19.48 \\
\hline
\end{tabular} 
\label{table:HOI_zeroshot_eval}
\vspace{-10pt}
\end{table}

\textbf{Unseen Composition}. Compared to FCL, our SEEM-based Seg2HOI framework shows improvements of 10.5\% and 8.8\% for rare-first and non-rare-first type, respectively, across full categories. Notably, for unseen categories, it achieves a margin of 6.9 mAP and 1.4 mAP over FCL. Unlike VCL, ATL, and FCL, which include specific losses for zero-shot HOI learning, our approach achieves superior performance without the need for such dedicated loss functions. This highlights the efficiency and effectiveness of our method in handling zero-shot HOI task. Compared to the baseline, we can confirm that the presence of masks provides a benefit for zero-shot HOI learning. While existing methods fine-tune the entire model, resulting in higher training costs, our approach shows significant performance improvements compared to VCL, where only the detector is frozen. Additionally, when comparing SEEM, which leverages CLIP embeddings, to MaskDINO, we observed a 23\% improvement in rare-first unseen categories under the rare-first setting. This performance is comparable to that of Gen-VLKT, which was fully fine-tuned on HICO-DET.

\begin{figure*}
	\centering
    \begin{subfigure}[b]{0.29\textwidth}
        \centering
        \includegraphics[width=\textwidth]{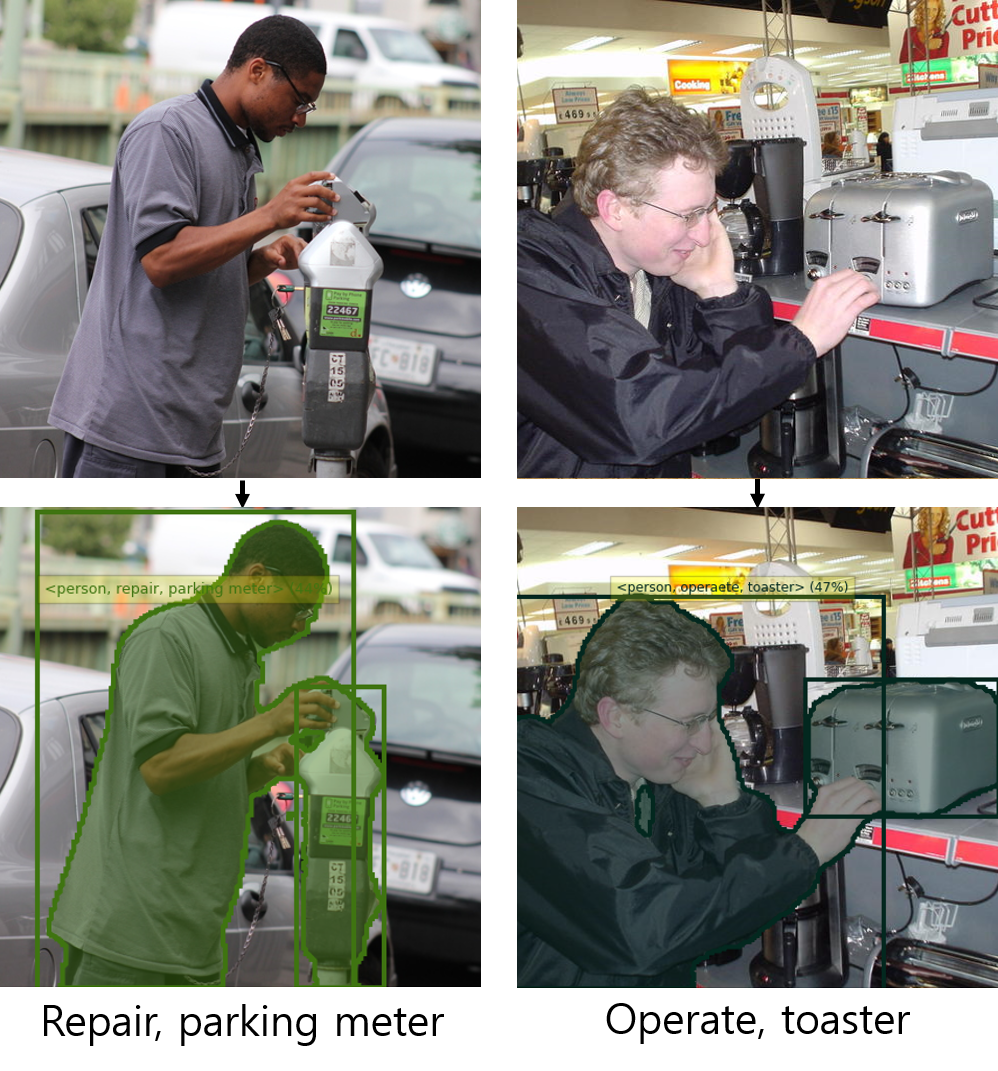}
        \caption{RF-UC}\label{fig:zero_shot_qualitative_RF}
    \end{subfigure} \hspace{0.005\linewidth}
    \begin{subfigure}[b]{0.365\textwidth}
        \centering
        \includegraphics[width=\textwidth]{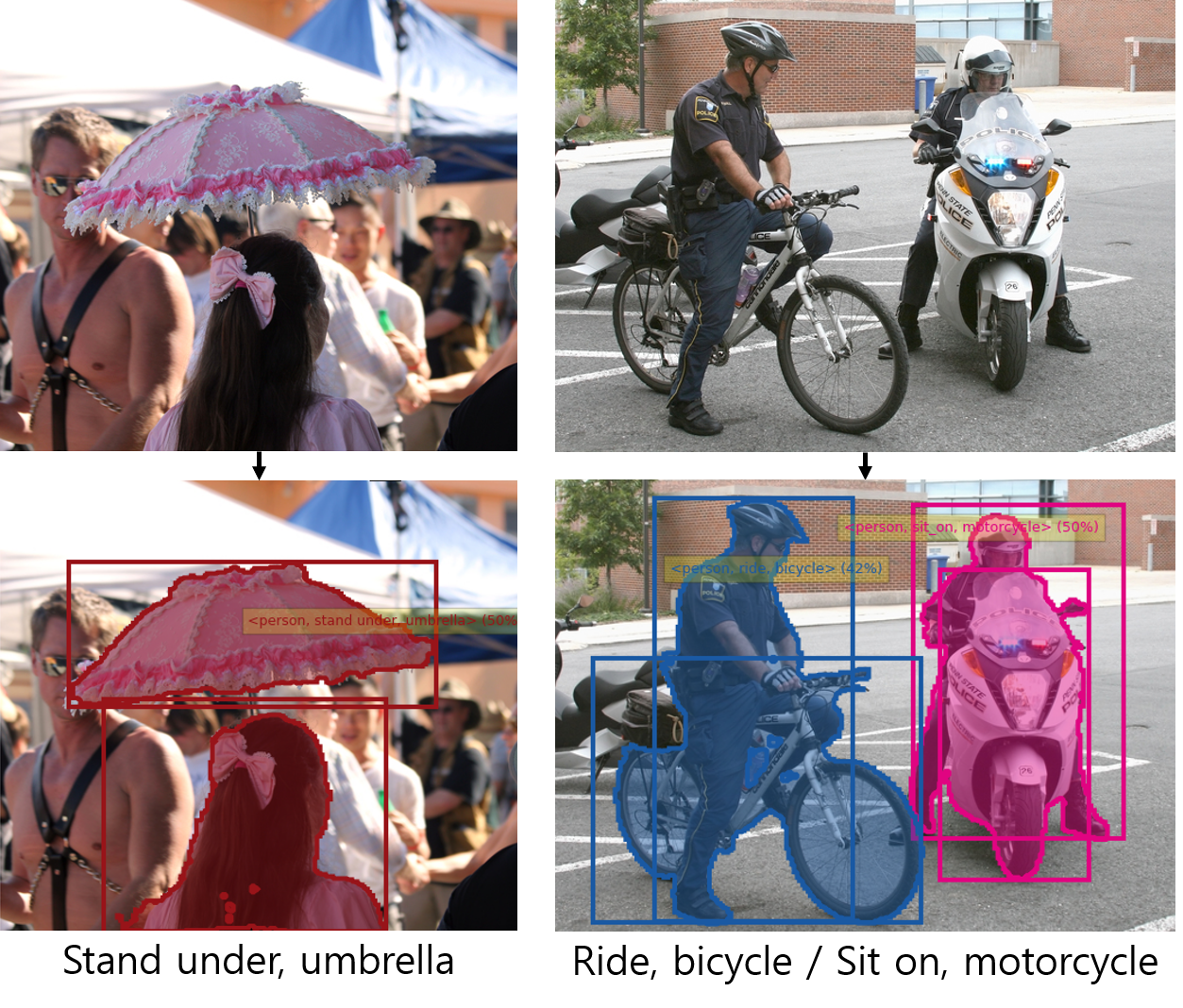}
        \caption{NF-UC}\label{fig:zero_shot_qualitative_NF}
    \end{subfigure}  \hspace{0.005\linewidth}
    \begin{subfigure}[b]{0.2925\textwidth}
        \centering
        \includegraphics[width=\textwidth]{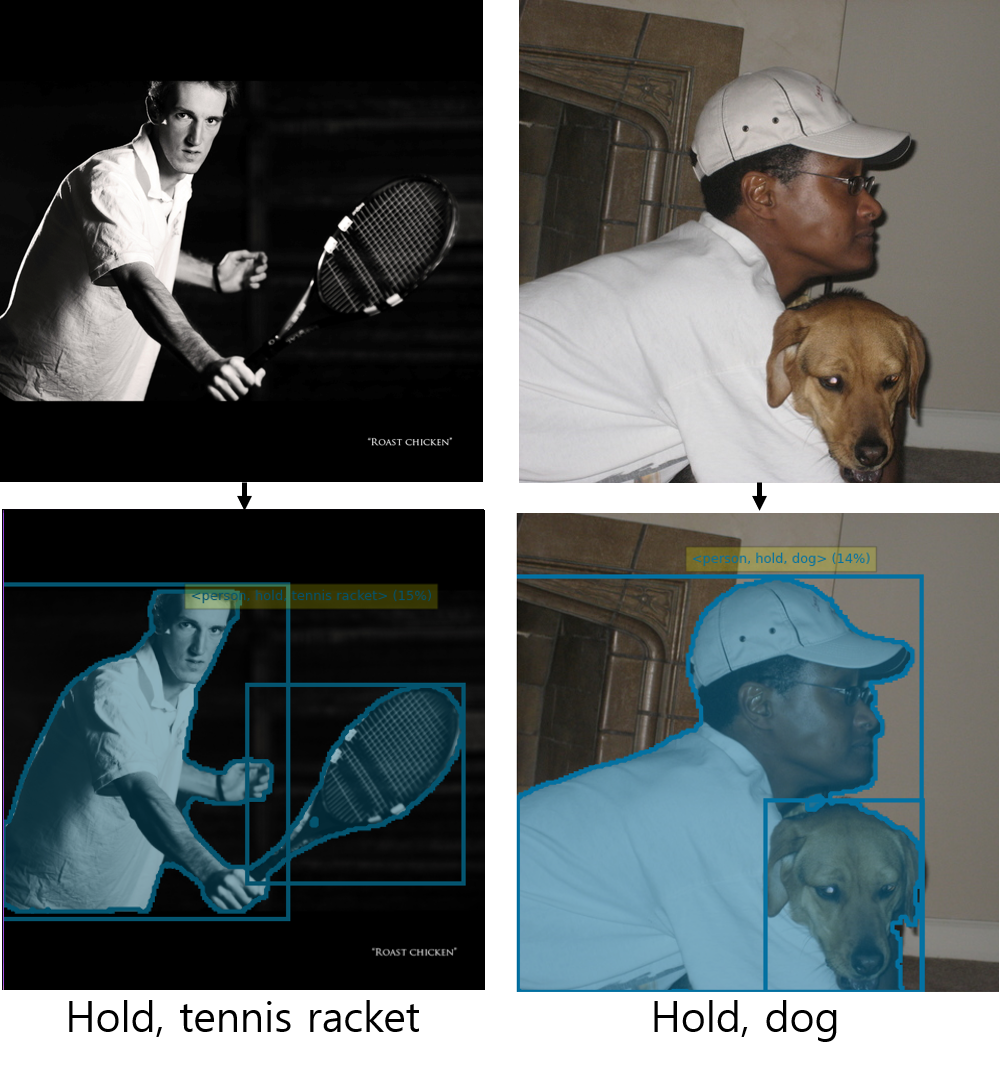}
        \caption{UO}\label{fig:zero_shot_qualitative_UO}
    \end{subfigure}
	\caption{{Qualitative results from zero-shot dataset. The sub-captions indicate the corresponding datasets.}}
	\label{figs:zero_shot_qualitative_seg}
    \vspace{-5pt}
\end{figure*}

\begin{figure*}
    \centering
    \begin{subfigure}[b]{0.484\textwidth}
        \centering
        \includegraphics[width=\textwidth]{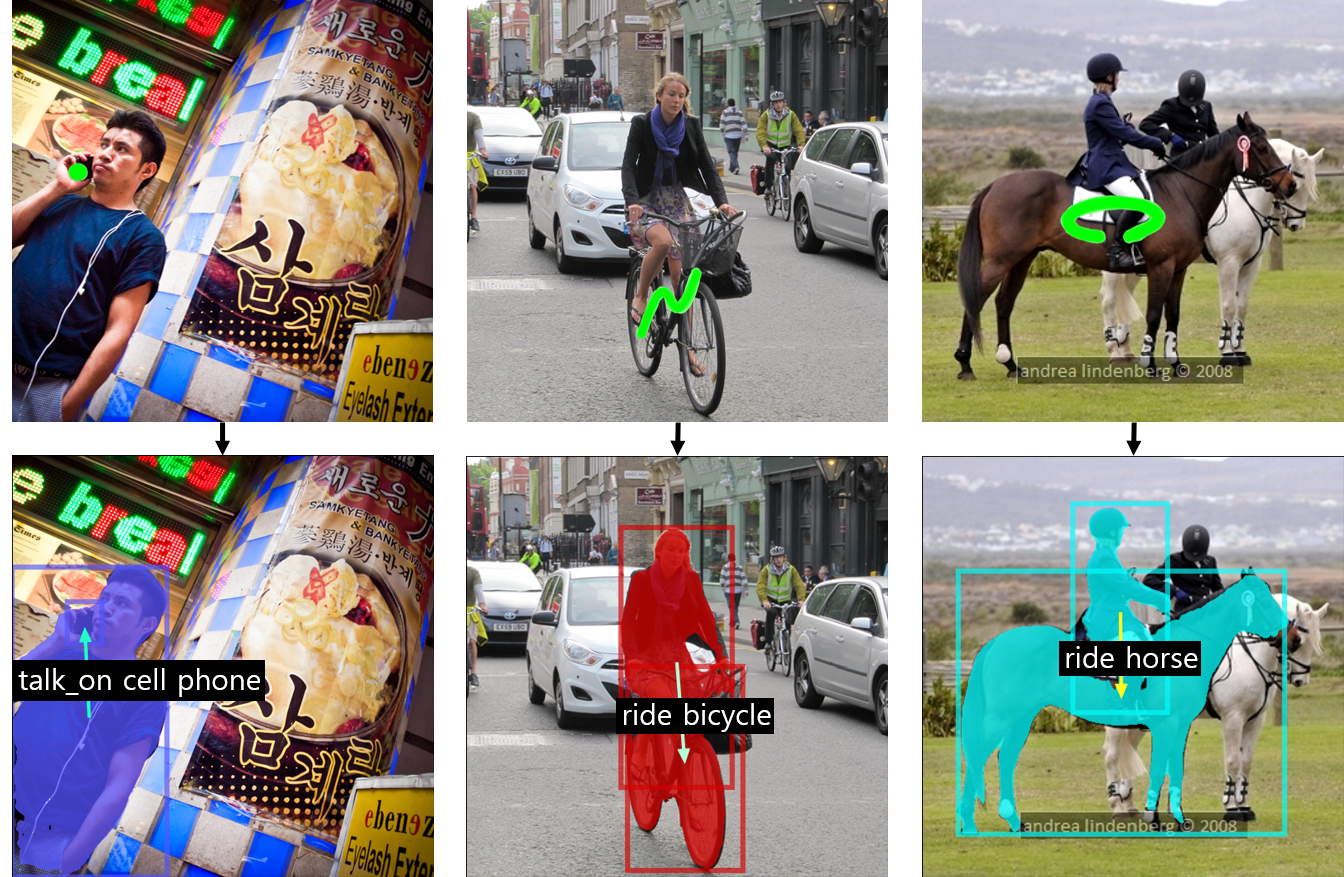}
        \caption{Visual prompt based HOI segmentation.}\label{fig:seem_qualitative_visual}
    \end{subfigure} \hspace{0.01\linewidth}
    \begin{subfigure}[b]{0.48\textwidth}
        \centering
        \includegraphics[width=\textwidth]{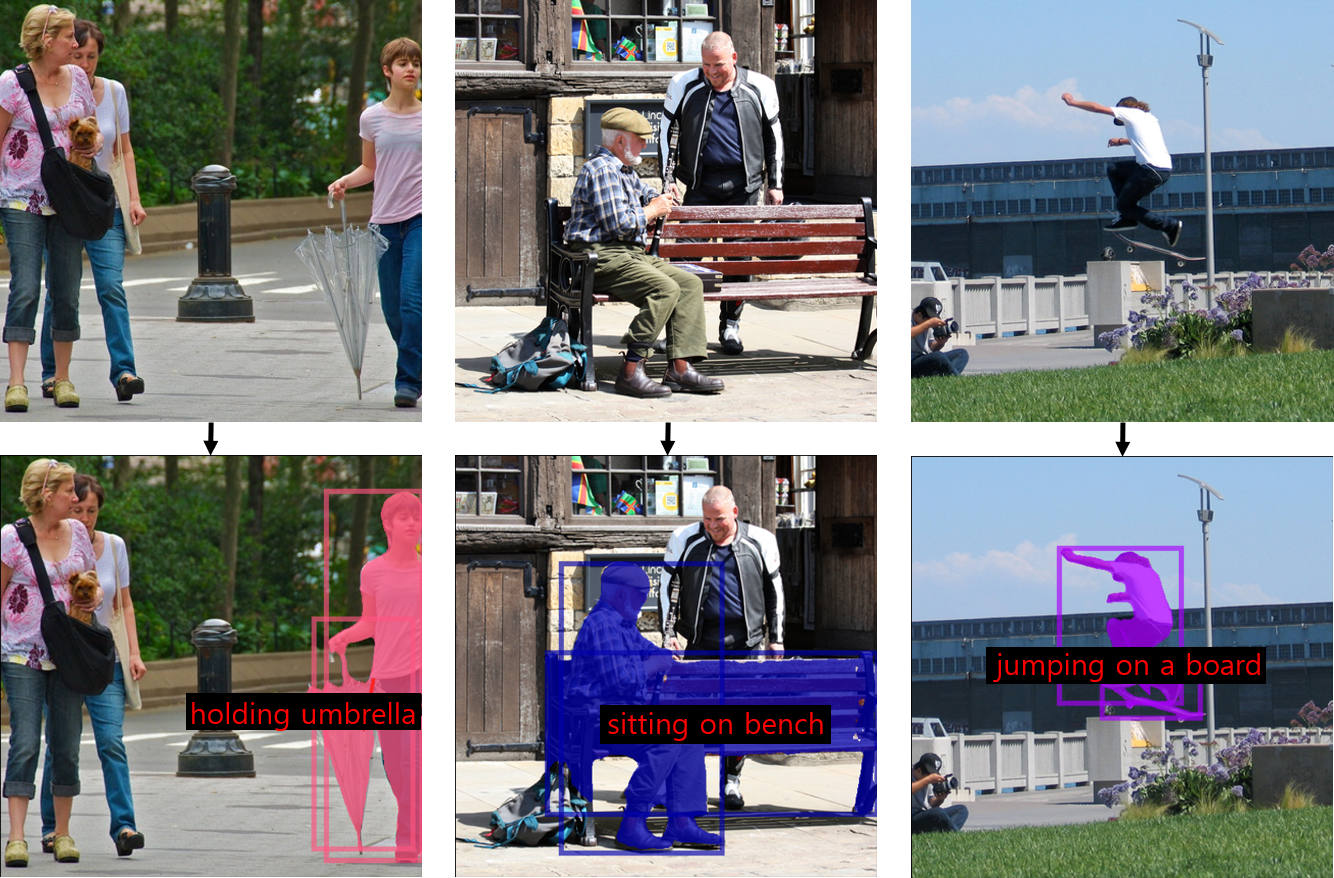}
        \caption{Text referring HOI segmentation.}\label{fig:seem_qualitative_text}
    \end{subfigure}
    \caption{Qualitative results using various prompts from Seg2HOI with SEEM. Our framework leverages the capabilities of foundation models, enabling its use across various applications. We present the quadruplets extracted from each prompt. The red text represents the user referred text prompt, while the white text on the image indicates the prediction with the highest similarity to the prompt.}
    \label{fig:seem2hoi_qualitative_result}
    \vspace{-15pt}
\end{figure*}

\textbf{Unseen Object}. Unlike our SEEM model, MaskDINO does not support open-vocabulary, making it impossible to experiment with unseen objects. Therefore, the experiments were conducted only on SEEM-based. Our framework shows significant performance improvements on unseen categories compared to ATL and FCL, and achieves performance comparable to Gen-VLKT, which also utilizes CLIP, confirming the effectiveness of CLIP text embeddings. Additionally, we observe a 89.9\% improvement over the baseline, demonstrating the impact of HOI masks even in experiments involving unseen object types. Compared to the ATL method, which freeze the backbone, our approach also shows a 81.6\% improvement on unseen categories.

\textbf{Quadruplet Qualitative Results.} We demonstrate that the model trained on the UC and UO datasets performs well on zero-shot HOI compositions. 
In Fig.~\ref{figs:zero_shot_qualitative_seg}, it can be observed that the model is capable of zero-shot detection for unseen HOIs. 
In Fig.~\ref{fig:zero_shot_qualitative_RF} and \ref{fig:zero_shot_qualitative_NF}, we observe that the model accurately predicts unseen HOI compositions. In Fig.~\ref{fig:zero_shot_qualitative_UO}, it demonstrates the ability to predict unseen objects and their corresponding action classes. In particular, we observe that the segmentation masks obtained are as clean and of similar quality to the HOI masks generated in the closed set.

\subsection{Prompt-Based Interactive Quadruplet Segmentation}\label{quadruplet_interactive_hoi_segmentation_experiments}
In this part, we present qualitative results demonstrating the output of quadruplets from visual and text prompts.
Fig.~\ref{fig:seem2hoi_qualitative_result} shows the results of the model combining the SEEM and Seg2HOI frameworks. 
In Fig.~\ref{fig:seem_qualitative_visual}, the model operates based on visual prompts, utilizing point and strokes. By applying a point or stroke on a specific object, we can obtain results where the interacting human is simultaneously segmented. 
In Fig.~\ref{fig:seem_qualitative_text},  the model operates based on text prompts, and we demonstrate this by inputting text prompts corresponding to the ground truth triplet labels. We can observe reasonable quadruplet results for both visual and text prompts.

\begin{figure*}
	\centering
    \begin{subfigure}[b]{0.315\textwidth}
        \centering
        \includegraphics[width=\textwidth]{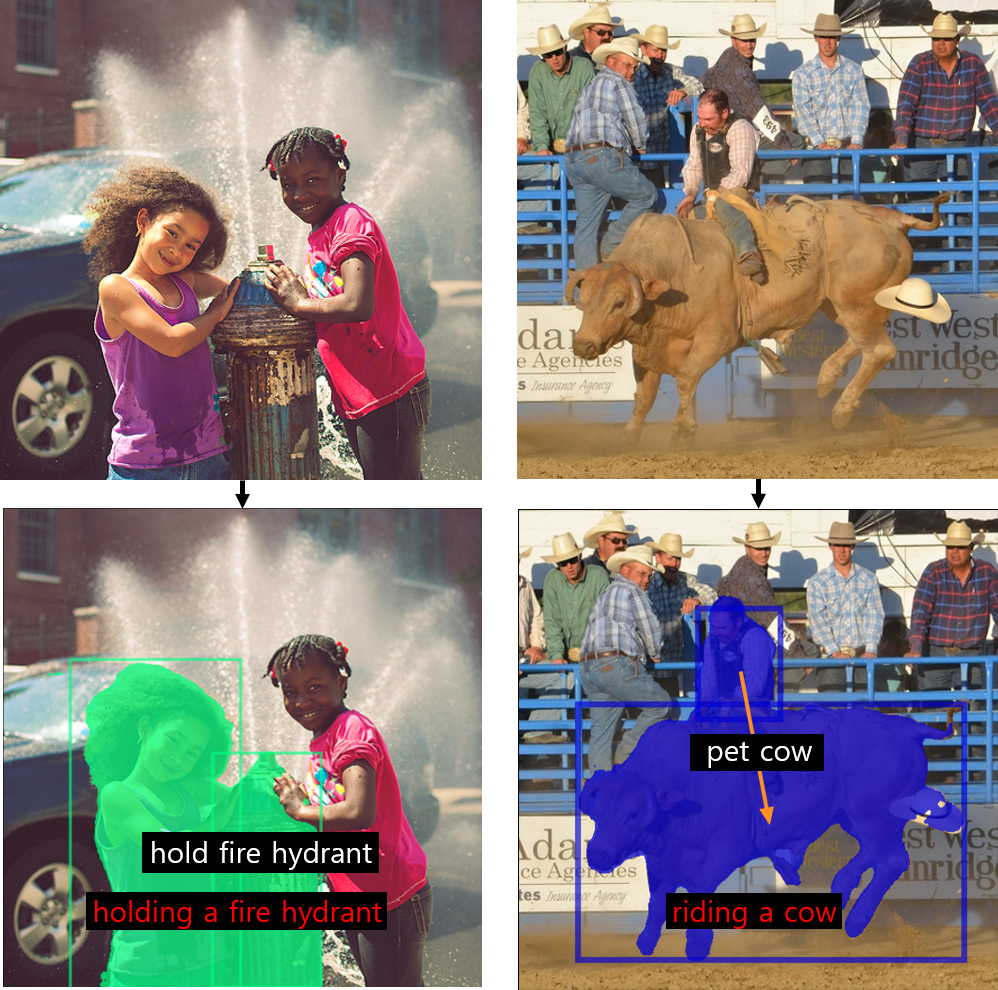}
        \caption{RF-UC}\label{fig:seem_qualitative_visual_RF}
    \end{subfigure} \hspace{0.005\linewidth}
    \begin{subfigure}[b]{0.315\textwidth}
        \centering
        \includegraphics[width=\textwidth]{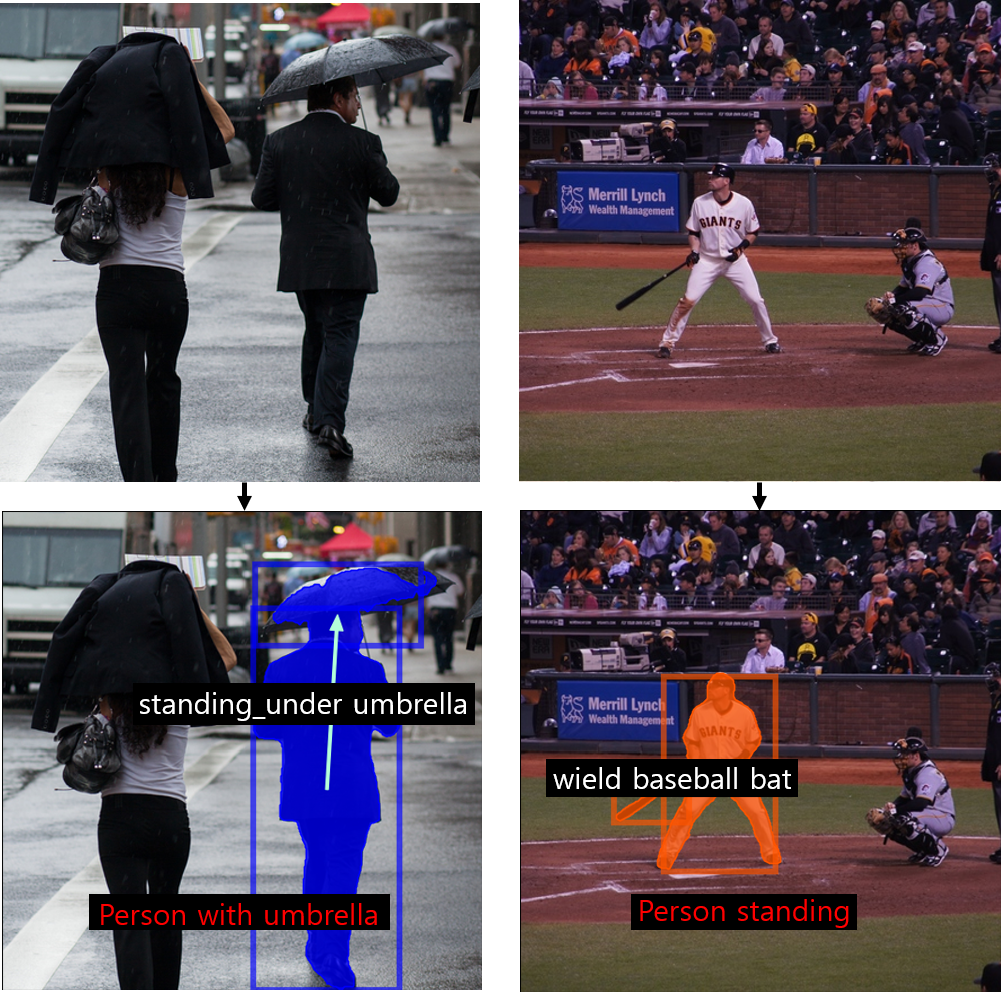}
        \caption{NF-UC}\label{fig:seem_qualitative_text_NF}
    \end{subfigure}  \hspace{0.005\linewidth}
    \begin{subfigure}[b]{0.313\textwidth}
        \centering
        \includegraphics[width=\textwidth]{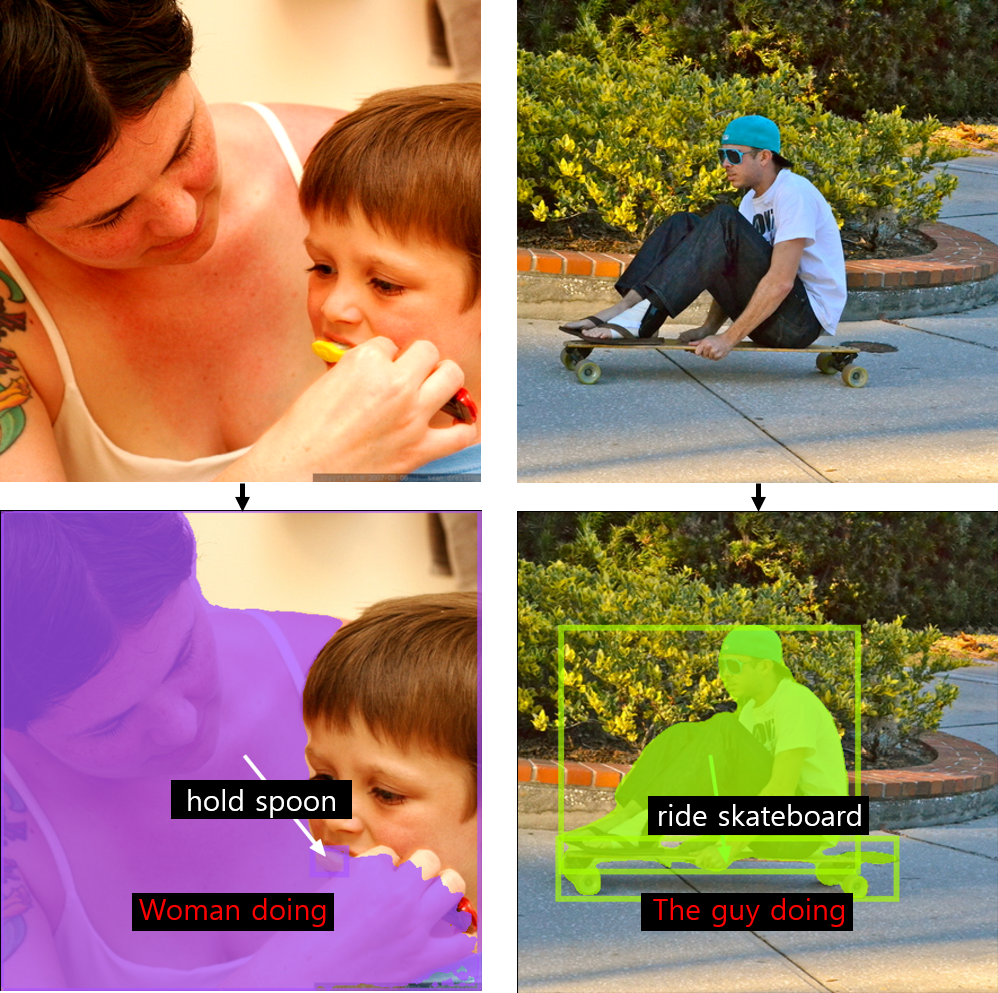}
        \caption{UO}\label{fig:seem_qualitative_text_UO}
    \end{subfigure}
	\caption{{Illustration of interactive HOI image segmentation in zero-shot datasets. The referred text can be seen on the masks in red. The sub-captions indicate the corresponding datasets.}}
	\label{figs:zero_shot_interactive_seg}
    \vspace{-5pt}
\end{figure*}

In Fig.~\ref{figs:zero_shot_interactive_seg}, an interactive demo is presented using a model trained on zero-shot data.
This demonstrates that HOI segmentation can be performed from the user text prompt, and the model produces effective outputs even with different word combinations from those used in the text embedding classifier during training. 
Especially Fig.~\ref{fig:seem_qualitative_visual_RF} shows that by entering the unseen composition's verb and object from the RF-UO data, the demo functions well, with the class having the highest similarity to the referred text being reasonably outputted. 
Fig.~\ref{fig:seem_qualitative_text_NF} demonstrates that even with more abstract prompts, the model still produces reasonable results corresponding to the referred text within the image. Lastly, Fig.~\ref{fig:seem_qualitative_text_UO} demonstrates the model's ability to handle unseen objects, where prompts like "Woman doing" and "The guy doing" result in HOI outcomes showing the woman and man interacting with objects in the image.
However, in the first column of Fig.~\ref{fig:seem_qualitative_text_UO}, an incorrect object is predicted, highlighting the model's lower performance on unseen objects and the need for future research to improve this aspect.

\subsection{Ablation Study}
In this section, we conducted several experiments to demonstrate the efficiency of our proposed framework. The experiments primarily focused on the V-COCO dataset, with additional experiments conducted on the HICO-DET dataset. 
For the ablation study, we conducted experiments by attaching our framework to the Mask-DINO ResNet-50 model.

\textbf{Effect of HOI Segmentation Mask}. 
In this part, we aim to verify the effectiveness of the proposed quadruplet. We assess the impact of the presence or absence of the HOI union mask $m^U$ and intersection mask $m^I$ to evaluate the effectiveness of the proposed methodology. 
As shown in Table~\ref{table:mask_ablation_hico}, we designate the model without the HOI mask as the baseline, which demonstrates the lowest performance across both the HICO and V-COCO datasets. 
On the other hand, using only HOI union masks $m^U$ achieved the highest performance on the V-COCO dataset, showing improvements of 2.0 mAP and 1.8 mAP over the baseline in Scenario 1 and Scenario 2, respectively. 
Furthermore, in the setting where the union mask $m^U$ is applied, the effectiveness of the HOI segmentation mask is evident across all datasets.
Even using the intersection mask alone improves performance over the baseline across all datasets. 
However, when the intersection area $b^I$ is small or no intersection area, pseudo labels cannot be generated, leading to noisy performance when data is insufficient.
This results in lower performance than the baseline in the HICO-DET Rare composition.

\begin{table}[ht]
\centering
\caption{Performance Comparison Based on the Presence of HOI segmentation mask. The results were obtained using the Seg2HOI model with Mask-DINO ResNet-50.}
\begin{tabular}{ccccccc}
\toprule
\multirow{2}{*}{$m_U$} & \multirow{2}{*}{$m_I$} & \multicolumn{3}{c}{HICO-DET (Default)} & \multicolumn{2}{c}{V-COCO} \\ \cmidrule(lr){3-5} \cmidrule(lr){6-7}
 & & \text{Full} & \text{Rare} & Non-Rare & $AP_\text{role}^{S1}$ & $AP_\text{role}^{S2}$ \\
\midrule
- & - & 28.33 & 23.53 & 29.76 & 64.8 & 67.1 \\
\checkmark & - & \textbf{29.67} & 24.38 & \textbf{31.24} & 66.8 & 68.9 \\
- & \checkmark & 28.94 & 22.44 & 30.88 & 65.3 & 67.6 \\
\checkmark & \checkmark & 29.55 & \textbf{24.49} & 31.06 & \textbf{67.3} & \textbf{69.2} \\
\bottomrule
\end{tabular}
\label{table:mask_ablation_hico}
\end{table}

\begin{table}[ht]
\centering
\caption{Performance comparison with different layer numbers of HOI decoder in our framework.}
\begin{tabular}{cccc}
\toprule
{\text{\# of layers ($L$)}} & $AP_\text{role}^{S1}$ & $AP_\text{role}^{S2}$ \\ 
\midrule
3 & 65.0 & 67.4 \\
6 & 66.8 & 68.9 \\
9 & \textbf{67.1} & \textbf{69.3} \\
\bottomrule
\end{tabular}
\label{table:layers_comparison}
\vspace{-5pt}
\end{table}

\textbf{Effect of Layer Number $L$ in Our Framework}.  
To explore the effectiveness of the Seg2HOI framework based on the number of layers, we conducted experiments by stacking 3 to 9 layers, as shown in Table~\ref{table:layers_comparison}. The best performance was achieved with 9 layers, providing a slight margin of 0.3 AP and 0.4 AP in scenario 1 and scenario 2, respectively, compared to the 6-layer setting.
However, although performance improves as the number of layers increases, considering the increase in parameters, we prioritize efficiency over marginal performance gains for practical applications. Therefore, in this paper, we report the 6 stacked layer setting as the optimal configuration for the Seg2HOI framework.

\section{Conclusion}
We present Seg2HOI, which expands the HOI downstream task from the vision foundation model in an implicit manner. Our methodology not only proposes HOI triplets but also quadruplets, demonstrating its effectiveness in improving model performance and its potential for various applications. Our framework freezes the vision foundation model, allowing it to transfer its capabilities to downstream tasks, enabling interactive segmentation from visual or text prompts without additional effort during training. Our methodology demonstrates comparable performance to existing HOI methods and achieves State-of-the-Art results on the HICO dataset while keeping the backbone frozen. Additionally, through various qualitative results, we confirm that our model produces robust outcomes and effectively operates even on zero-shot data. We believe that our work offers an efficient approach to scene understanding tasks, building upon foundation models designed for fundamental tasks like detection and segmentation. As a first step towards segmentation-based HOI, we aim for it to inspire further active research in this field.












\printcredits

\section*{Declaration of competing interest}
The authors declare that they have no known competing financial interests or personal relationships that could have appeared to influence the work reported in this paper.

\section*{Data availability}
The data used in this study are available at the following repositories: \url{https://github.com/s-gupta/v-coco} and \url{https://github.com/fredzzhang/hicodet}.

\section*{Acknowledgements}
This work was supported by Institute of Information \& communications Technology Planning \& Evaluation (IITP) grant funded by the Korea government (MSIT) (No. 2022-0-00871, Development of AI Autonomy and Knowledge Enhancement for AI Agent Collaboration), and the National Research Foundation of Korea (NRF) grant funded by the Korea government (MSIT) (No. RS-2023-00211357)

\bibliographystyle{cas-model2-names}




\end{document}